\documentclass[10pt,letterpaper]{article}
\usepackage{cogsci}
\cogscifinalcopy 

\usepackage[T1]{fontenc}
\usepackage{microtype}
\usepackage{float}
\usepackage{listings}
\usepackage{float}
\usepackage{amsmath}
\usepackage[skins,breakable,listings]{tcolorbox}
\usepackage{graphicx}
\usepackage{subcaption}
\usepackage{url}
\usepackage{appendix}

\usepackage[
  style=apa,
  natbib=true,
  hyperref=true,
  annotation=false,
]{biblatex}
\DeclareCiteCommand{\citep}[\mkbibparens]
  {\usebibmacro{prenote}}
  {\usebibmacro{citeindex}%
   \printtext[bibhyperref]{\usebibmacro{cite}}}
  {\multicitedelim}
  {\usebibmacro{postnote}}
  
\usepackage{enumitem}
\setlist[itemize]{leftmargin=2em, topsep=0pt}
\setlist[enumerate]{leftmargin=2em, topsep=0pt}

\usepackage{multirow}
\usepackage{booktabs}
\usepackage{titlesec}
\titlespacing*{\paragraph}
{0pt}        
{0.3ex}      
{0.5em}      

\newcommand{\eg}{e.g.\,}
\definecolor{myred}{rgb}{0.8,0,0}
\definecolor{background}{rgb}{0.92941176, 0.96078431, 0.9686}
\definecolor{line}{rgb}{0.27058824, 0.38823529, 0.419}
\newcommand{\todo}[1]{}
\renewcommand{\todo}[1]{{\color{myred} Todo: {#1}}}
\definecolor{myblue}{rgb}{0.40,0.25,0.60}
\usepackage{hyperref}
\hypersetup{
  colorlinks=true,
  linkcolor=myblue,
  citecolor=myblue,
  urlcolor=myblue,
  pdfproducer={}
}

\makeatletter
\let\latex@showhyphens\showhyphens
\makeatother

\newtcblisting{pythonbox}{
    listing only,
    listing options={
        language=Python,
        basicstyle=\ttfamily\fontsize{6}{8}\selectfont,
        keywordstyle=\color{blue},
        commentstyle=\color{line},
        stringstyle=\color{green!50!black},
        showstringspaces=false
    },
    colback=background,
    colframe=line,
    arc=3pt,
    boxrule=2pt,
    left=3pt,
    right=3pt,
    top=0pt,
    bottom=0pt
}

\lstdefinelanguage{none}{}
\newtcblisting{tcbverbatim}[2][]{ 
  blank,
  borderline={1pt}{-2pt},
  listing only,
  breakable,
  enhanced jigsaw,
  colback=white,
  left=10pt,
  toptitle=5pt,
  title={#1},
  fonttitle=\sffamily\bfseries,
  coltitle=black,
  colbacktitle=gray!10,
  listing options={
    language=none,
    escapeinside={(*@}{@*)}, 
    basicstyle=\ttfamily\small\parindent=0pt,
    columns=flexible,
    breaklines=true,
    breakatwhitespace=true,
    breakautoindent=false,
    breakindent=0pt,
    #2 
  }
}

\title{Discovering Adaptive Transmission Programs for Collective Innovation}

\author[1]{Cédric Colas}
\author[1]{Jérémy Perez}
\author[2]{Eleni Nisioti}
\author[3]{Akhilesh Mocherla}
\author[1]{Pierre-Yves Oudeyer}
\author[1]{Clément Moulin-Frier}
\author[4]{Maxime Derex}

\affil[1]{Flowers AI \& CogSci Lab, Inria, France}
\affil[2]{IT University of Copenhagen, Denmark}
\affil[3]{Independent Researcher, UK}
\affil[4]{Institute for Advanced Study in Toulouse, France}

\begin{document}

\maketitle

\begin{abstract}
Human collective intelligence depends on transmission processes: who shares what with whom, how, and when. While these processes emerge from individual cognition, they can also be directed by deliberate top-down protocols. Prior work has studied how transmission shapes collective outcomes primarily through the lens of network structure, varying who shares with whom and when. But networks are state-agnostic: they cannot condition transmission on what agents know or on the state of the collective. Here, we formalize transmission protocols as state-aware programs that route information and resources based on agent and collective states, and we use LLM-guided evolutionary search to design effective protocols in a collective discovery task. Evolved protocols increase collective performance over standard baselines from the literature by up to 37\%. Ablations confirm that state-awareness drives this advantage: removing content-dependence while preserving network topology and timing eliminates performance gains. We find that evolved protocols also transfer across domain variations and agent populations. These results demonstrate that effective and generalizable transmission protocols can be discovered in silico, suggesting a path toward AI-assisted design of coordination infrastructure that enhances human collective intelligence.

\vspace{-1.5mm}
\textbf{Code and Supplementary} \href{https://tinyurl.com/protocol-discovery}{tinyurl.com/protocol-discovery}

\vspace{-1.5mm}

\end{abstract}

\setlength{\parindent}{0pt}
\setlength{\parskip}{6pt}
\section{Introduction}
Human intelligence is collective. Our species' achievements, from cumulative technology to scientific knowledge and complex institutions, arise not from individual brilliance but from our capacity to accumulate and recombine knowledge across individuals and generations~\citep{boyd1988culture, henrich2015secret, mesoudi2018cumulative, falandays2023all}. How this knowledge flows\,---\,who shares what with whom, how, and when\,---\,shapes whether groups solve problems efficiently or squander their collective potential~\citep{woolley2010evidence, derex_partial_2016, riedl2021quantifying}.

Transmission processes are shaped by both bottom-up and top-down forces. They emerge bottom-up from human social cognition: who we attend to, what we share, how we imitate or teach, and when~\citep{henrich2001evolution, rendell2010copy, kendal2018social, kline2015learn, schultner2025feature}. But they can also be directed top-down by deliberate \textit{transmission protocols}: \eg the reporting structures of organizations, the publication norms of scientific communities, or the information routing of recommendation systems. Unlike bottom-up processes, protocols can be purposefully designed, making them a natural lever for enhancing human collective intelligence. Realizing this potential requires understanding which aspects of transmission matter for collective performance, and developing novel methods to discover effective protocols.

\begin{figure*}[ht]
    \centering
    \includegraphics[width=\linewidth]{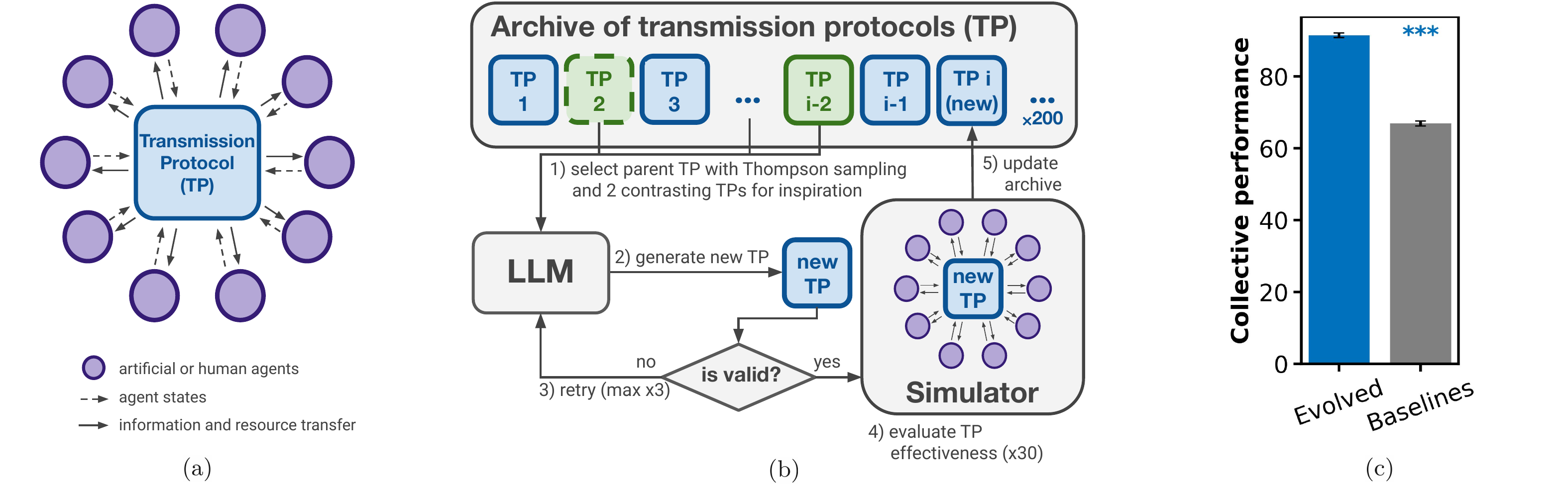}
    \caption{\textbf{Discovering adaptive transmission protocols in silico}. (a) State-aware transmission protocols (TPs) route information and resources between agents based on agent and collective states. (b) We search for effective TPs with an evolutionary algorithm that leverages large language models (LLMs) to generate and refine candidate protocols expressed as stateful programs. (c) Best evolved protocols increase collective performance in simulation by 37\% over baselines (mean$\pm$sem, N=5, p<0.001). \label{fig:main}}
\end{figure*}

\paragraph{Understanding how transmission shapes collective outcomes.}
Prior work has investigated transmission primarily through the lens of network structure~\citep{centola2022network}. Most studies vary \textit{who shares with whom} through network topology, consistently finding that intermediate connectivity outperforms both isolation and full connectivity: dense networks cause premature convergence, while sparser structures preserve the diversity needed to discover complex solutions~\citep{lazer2007network, barkoczi2016social, migliano2020hunter, cantor2021social, garg2025causal}. Other studies also vary \textit{when} sharing occurs through simple dynamic network structures, such as periodic visits between otherwise isolated pairs~\citep{derex_partial_2016, nisioti_social_2022, nisioti2024collective} or alternating phases of independent and collaborative work~\citep{bernstein2018intermittent}, generally finding that intermittent collaboration outperforms continuous sharing.

Yet the framing of transmission as network structure is fundamentally limited. Networks are state-agnostic: transmission decisions do not depend on what agents know or on the state of the collective as a whole. Real-life transmission processes, by contrast, are deeply state-dependent. People selectively attend to experts, share what is relevant, and seek information when needed~\citep{kendal2018social, henrich2001evolution} and top-down systems like recommendation algorithms and organizational hierarchies similarly condition routing on context. Recent work points to the same gap: finding that group synergy\,---\,whether agents hold complementary rather than redundant knowledge\,---\,predicts collective performance better than network topology alone~\citep{garg2025causal}. Yet networks cannot foster such synergy because they have no way to shape what is shared, with whom, or when, based on what agents currently know. Expanding the design space to allow protocols that condition transmission on agent and collective states could yield far more adaptive coordination, and with it, larger potential gains from well-designed interventions.

\paragraph{Designing interventions to improve collective outcomes.}
Designing such interventions is a central problem in mechanism design\,---\,a field studying how rules and institutions can be crafted to align individual behavior with group-level goals~\citep{hurwicz2006designing, maskin2008mechanism}. In the collective intelligence literature, interventions have so far been designed by hand, targeting specific known biases or communication breakdowns~\citep{jayles2021crowd, tessler2024ai}. In economics, deep mechanism design automates this process using reinforcement learning to discover effective interventions in simulation~\citep{zheng2022ai, koster2022human, tacchetti2025deep}. The resulting policies are black-box neural networks, and thus offer limited insight into why a specific intervention works or whether it might generalize beyond the specific setting in which it was trained.

\paragraph{Our approach.}
We take a different approach: formalizing transmission protocols as interpretable, state-aware programs, searching for effective protocols in silico via program-search, and systematically evaluating whether discovered protocols generalize across agent populations and task configurations.

We study collective innovation in \textit{Collective Little Alchemy}, an adaptation of the potion model~\citep{derex_partial_2016, migliano2020hunter, cantor2021social, garg2025causal, nisioti_social_2022} based on the online game Little Alchemy 2.\footnote{https://littlealchemy2.com/} Agents combine elements to discover new ones (\eg \textit{water + fire} $\rightarrow$ \textit{steam}), which can then be used in further combinations. A transmission protocol determines which memories of past combinations are shared between agents and when, temporarily making the involved elements available to recipients. The domain features cumulative discovery and path dependencies, making collective performance\,---\,the total unique elements discovered by the group\,---\,sensitive to how information and resources are routed.

To move beyond the state-agnostic network paradigm, we formalize transmission protocols as stateful executable programs that, for each timestep, take as input agent states (memories and inventories) and specify who shares what with whom (Figure~\ref{fig:main}a). We search over this design space using an evolutionary algorithm that proposes candidate protocols as code, evaluates their effectiveness via simulation, and iteratively refines high-performing candidates (Figure~\ref{fig:main}b). Candidate generation and refinement rely on a language model, as the design space is too unconstrained for hand-crafted program generators~\citep{lehman2023evolution}. To assess whether discovered protocols generalize, we evaluate them on domain configurations and agent populations they were not optimized for.

\paragraph{Findings.}
Discovered protocols outperform baselines from the literature by up to 37\% (Figure~\ref{fig:main}c). Ablations confirm that this advantage depends on state-awareness: removing state-dependent routing while preserving network topology and timing eliminates performance gains. Protocols also generalize across agent types and domain configurations they were not optimized for, suggesting they adapt to local conditions rather than overfitting to the training configuration. These results demonstrate that in silico search can discover effective and generalizable transmission protocols, suggesting a path toward AI-assisted design of coordination infrastructure that enhances human collective intelligence~\citep{cui2024ai}.

\section{Methods}
\label{sec:methods}
We develop a simulation framework for the study of collective innovation based on the discovery game Little Alchemy 2 , populate it with agents that approximate human decision-making , and formalize transmission protocols as state-aware programs. We then use an LLM-guided evolutionary algorithm to search for effective protocols, and evaluate whether discovered protocols leverage state-awareness and generalize across domain variants and agent populations. Appendices can be found in the Supplementary pdf at \href{https://tinyurl.com/protocol-discovery}{tinyurl.com/protocol-discovery}.

\subsection{Collective Little Alchemy}
\label{sec:domain}

We study collective innovation with a \textit{potion model}: an experimental paradigm in which participants combine elements to discover new ones, which can then be used in further combinations~\citep{derex_partial_2016, cantor2021social, migliano2020hunter, nisioti_social_2022}. This task structure captures core features of real-world innovation: discovery is cumulative, different exploration paths lead to different outcomes, and sharing across paths can unlock combinations that no single path would reach. Inspired by earlier work on artificial collective intelligence~\citep{nisioti2024collective}, we introduce the \textit{Collective Little Alchemy} task, based on the browser game Little Alchemy 2, which features 720 elements with deterministic combination rules reflecting real-world relationships (\eg water + fire $\rightarrow$ steam). 

A population of $N$ agents explores the game over $T$ synchronous steps. Each agent maintains an inventory of discovered elements, initialized with four base elements (water, fire, earth, air), and a memory of past combinations represented as triples $(e_1, e_2, r)$ where $r$ is the resulting element for valid combinations, or indicates failure otherwise. At each step, agents simultaneously attempt a combination of elements from their inventories. A transmission protocol then determines which memories are shared between agents. When an agent receives a memory, the elements involved in the combination, including any result, become available in a temporary social inventory that persists only for the current step. Agents may use social elements in their next combination, but only elements discovered through their own successful combinations permanently enter their inventory. We measure collective performance as the total number of unique elements discovered by at least one agent, analogous to patent counts or publication output as measures of collective innovation~\citep[\eg][]{posch2024social}. Because only self-discovered elements enter an agent's inventory, each agent can discover at most one new element per step. This ensures that collective performance reflects how information and resources are routed, not simply how much is transmitted.


\subsection{Agent models}
\label{sec:agents}

The effectiveness of a transmission protocol may depend on the agents it organizes. As our primary agents, we adopt the empowerment-maximizing model of Brändle et al., which best predicts human combination choices in Little Alchemy~\citep{brandle_empowerment_2023}.

\paragraph{Empowerment agents.} Agents select combinations via softmax (temperature $\beta=0.05$) over the empowerment of each possible action, discarding actions with zero empowerment. The empowerment of a combination is the estimated expected empowerment of the combination's outcome conditioned on agent's $i$ knowledge at game step $t$~\citep{brandle_empowerment_2023}:

{\setlength{\abovedisplayskip}{1pt}
 \setlength{\belowdisplayskip}{2pt}
\begin{align*}
\text{Emp}^\text{act}_i(e_1, e_2, t) &= P_i(\text{success} \mid e_1, e_2, t) \\
&\quad \cdot \sum_{e_r} P_i(e_r \mid e_1, e_2, t) \cdot \text{Emp}^\text{item}_i(e_r, t).
\end{align*}
}

This requires three components. First, the empowerment of obtaining an element $e$, $\text{Emp}^\text{item}_i(e, t)$: zero if the agent already owns $e$, otherwise the count of unique elements that $e$ could be used to produce. We use ground-truth counts as in the original model, adding multiplicative Gaussian noise ($\mu=1, \sigma=0.1$) for heterogeneity across agents at initialization. Second and third, the success probability $P_i(\text{success}\mid e_1, e_2, t)$ and the transition distribution $P_i(e_r \mid e_1, e_2, t)$, initialized by the offline training procedure of Brändle et al.\ (details in Appendix~\ref{app:agents}) and updated as agents accumulate knowledge. Failed combinations (from own attempts or social memories) become deterministic failures (empowerment zero); successful own attempts become deterministic with zero empowerment (the agent already owns the result); and successful social memories become deterministic with empowerment equal to that of the known outcome.

Finally, to model how humans incorporate social information \citep{fromhold-eisebithTracingSocialDimension2014, cialdiniSocialInfluenceCompliance2004}, we introduce a \emph{social bias}: when an agent's social inventory is non-empty, there is probability $p_{\text{social}}=0.5$ that it restricts its candidate set to combinations involving at least one socially received element. This fixes how agents process received transmissions, allowing us to isolate the effects of protocol-level routing decisions.

\paragraph{Alternative agents.} We implement two alternative agent models to test generalization. \emph{Stochastic agents} select uniformly among untried combinations, using the same social bias mechanism as empowerment agents. \emph{LLM agents} use a large language model (\texttt{Gemini 3.0 Flash}) prompted with their current inventory, socially available elements, individual and social memories to reason about which combinations to attempt (see full prompt in Appendix~\ref{app:prompts}). LLM agents incorporate social information through their own reasoning rather than the fixed social bias mechanism.

\paragraph{Attention bottleneck.} For all agent types, we impose an attention bottleneck: each agent receives at most 20 memories per step, subsampled uniformly when more are sent. This models bounded attention\,---\,human players could not realistically process hundreds of shared memories at each step\,---\,and ensures that protocol effectiveness depends on routing information strategically rather than simply broadcasting everything.


\subsection{Transmission protocols (TP)}
\label{sec:protocols}

A transmission protocol (TP) is a Python class that maintains internal state across the simulation and exposes a \texttt{share} method that, at each game step, observes agent states and outputs sharing decisions:
\begin{pythonbox}
TP = TransmissionProtocol(n_agents, n_steps)
exchanges = TP.share(i_step, agent_states) 
# agent_states: {agent_id: {inventory: {e_1, e_2, e_3, ...}, 
#                           memories: [(e_1, e_2, e_r), ...]}}
# exchanges: {recipient_id: [(sender_id, memory_id), ...]}
\end{pythonbox}

\paragraph{Baselines.} This formalization can encode standard approaches from the collective intelligence literature. We implement five baselines spanning from no sharing to dense sharing: \textit{Asocial} (no sharing), \textit{Paired} (fixed dyads exchange up to 10 random memories with 50\% probability per step), \textit{Dynamic} (same as Paired, but agents periodically visit other dyads~\citep{derex_partial_2016, nisioti_social_2022, nisioti2024collective}), \textit{Graph} (agents share along edges of an Erdős–Rényi network with $p=0.2$~\citep{lazer2007network, mason2008propagation}), and \textit{Stochastic} (each agent receives 20 randomly sampled memories per step from other agents, saturating the attention bottleneck).

\paragraph{Beyond networks.} Unlike network-based approaches, which fix who shares with whom and when independently of agent states, TPs can condition all routing dimensions on agent and collective states. A protocol might share only elements that the recipient lacks, route failed combinations to agents who hold both ingredients to prevent mistakes, prioritize recently discovered recipes, or reduce sharing as common items saturate the population. The following example illustrates a protocol that adapts sharing intensity to collective progress (increasing when the group makes discoveries, decaying during stagnation) and routes memories only to agents who lack the elements involved:
\begin{pythonbox}
class TransmissionProtocol:
  def __init__(self, n_agents, n_steps):
    self.rate, self.prev = 0.5, 0
    
  def share(self, i_step, agent_states):
    # update sharing rate
    all_inv = [s['inventory'] for s in agent_states.values()]
    n_new = len(set().union(*all_inv)) - self.prev
    self.prev = len(set().union(*all_inv))
    self.rate = min(1, self.rate + 0.1 * n_new) if n_new > 0 \
                else self.rate * 0.9
    # route memories
    exchanges = {aid: [] for aid in agent_states}
    for sender, state in agent_states.items():
      for memory in state['memories']:
        e1, e2, r = memory
        for recvr, r_st in agent_states.items():
          if recvr == sender: continue
          lacks = (e1 not in r_st['inventory']
                   or e2 not in r_st['inventory'])
          if lacks and random() < self.rate:
            exchanges[recvr].append((sender, memory))
    return exchanges
\end{pythonbox}

The design space is deliberately constrained: protocols observe agent states but cannot directly modify them and have no access to the ground-truth recipe graph. Because agents retain fully autonomy, differences in collective performance purely arise from information and resource routing.

\subsection{LLM-guided Evolutionary search}
\label{sec:evolution}

We use an LLM-guided evolutionary algorithm to search over the protocol design space, building on recent work showing that large language models can generate and refine programs through iterative variation and selection~\citep{lehman2023evolution, meyerson2024language, romera2024mathematical, pourcel2024aces, novikov2506alphaevolve, lange2025shinkaevolve, zhang2025darwin}. The algorithm maintains an archive of protocols, each represented as executable Python code together with its fitness score and a natural-language description. We seed the archive with the five baseline protocols.

The search proceeds in two phases over 200 iterations using the \texttt{Gemini 3.0 Flash} model. During a \emph{warmup phase} (first 50 iterations), the LLM generates novel protocols conditioned on the five baseline protocols, bootstrapping archive diversity. During the \emph{evolution phase}, each iteration selects a parent protocol via Thompson sampling over fitness, samples two contrast solutions with a soft fitness bias from outside the parent's lineage, and prompts the LLM to either \emph{improve} the parent or \emph{diversify} from the past solutions by proposing a fundamentally different strategy (details in Appendix~\ref{app:evolution} and prompt in Appendix~\ref{app:prompts}).

Each candidate is validated by executing it in a short sandboxed simulation (with up to three repair attempts if execution fails). The fitness of valid protocols (their effectiveness) is evaluated as the mean collective performance across 30 simulations with 10 empowerment agents over 150 game steps. All valid protocols enter the archive regardless of fitness, preserving diversity. We also run \emph{generation-only} processes that use the warmup procedure for the 200 iterations, producing statistically independent protocols that enable sound correlation analyses.


\subsection{Evaluation}
\label{sec:evaluation}

Our evaluation addresses two questions: do effective protocols leverage state-awareness, and do discovered protocols generalize to new settings?

\paragraph{State-awareness ablation.} To test whether state-aware content selection drives protocol effectiveness, we construct a controlled ablation that replays each protocol's exact transmission schedule\,---\,who shares with whom, how many memories, and at which step\,---\,but randomly selects which memories to send. This preserves network topology and timing while removing state-dependent content routing. We compare fitness between original and ablated protocols (details in Appendix~\ref{app:ablation}). We also compute a set of behavioral features from both original and ablated simulation logs across $1{,}000$ independently generated protocols (see list in Appendix~\ref{app:features}). For each feature, we subtract the ablated value from the original, yielding a per-protocol measure of how much state-aware routing shifts that feature relative to random selection. Correlating these shifts with fitness reveals which state-aware behaviors predict performance: a positive correlation means protocols that increase that feature (relative to random) tend to perform better.   

\paragraph{Generalization tests.} We evaluate discovered protocols on configurations they were not optimized for (training: 10 empowerment agents, 150 steps, 720 elements). Population variations test three axes: \emph{size} (6 and 20 agents), \emph{agent type} (stochastic agents, LLM agents, and mixed populations), and \emph{behavioral parameters} (stronger empowerment bias $\beta/2$ or stronger social bias $p_{social}=0.9$). Domain variations alter the game structure: \emph{depth-limited} restricts to elements within 3 combination steps of base elements (86 reachable), \emph{3-item starts} initializes agents with 3 of 4 base elements (587--684 reachable), \emph{heterogeneous starts} assigns each agent a different pair of base elements, and \emph{item dropout} randomly removes elements until fewer than 300 remain reachable. Each configuration is evaluated over 30 simulations.

\section{Results}
\label{sec:results}

We first assess whether LLM-guided search discovers protocols that outperform baselines, then test whether their advantage depends on state-awareness, and finally evaluate generalization across populations and domains.

\begin{figure}[ht]
    \centering
    \begin{subfigure}{0.49\linewidth}
        \centering
        \includegraphics[width=\linewidth]{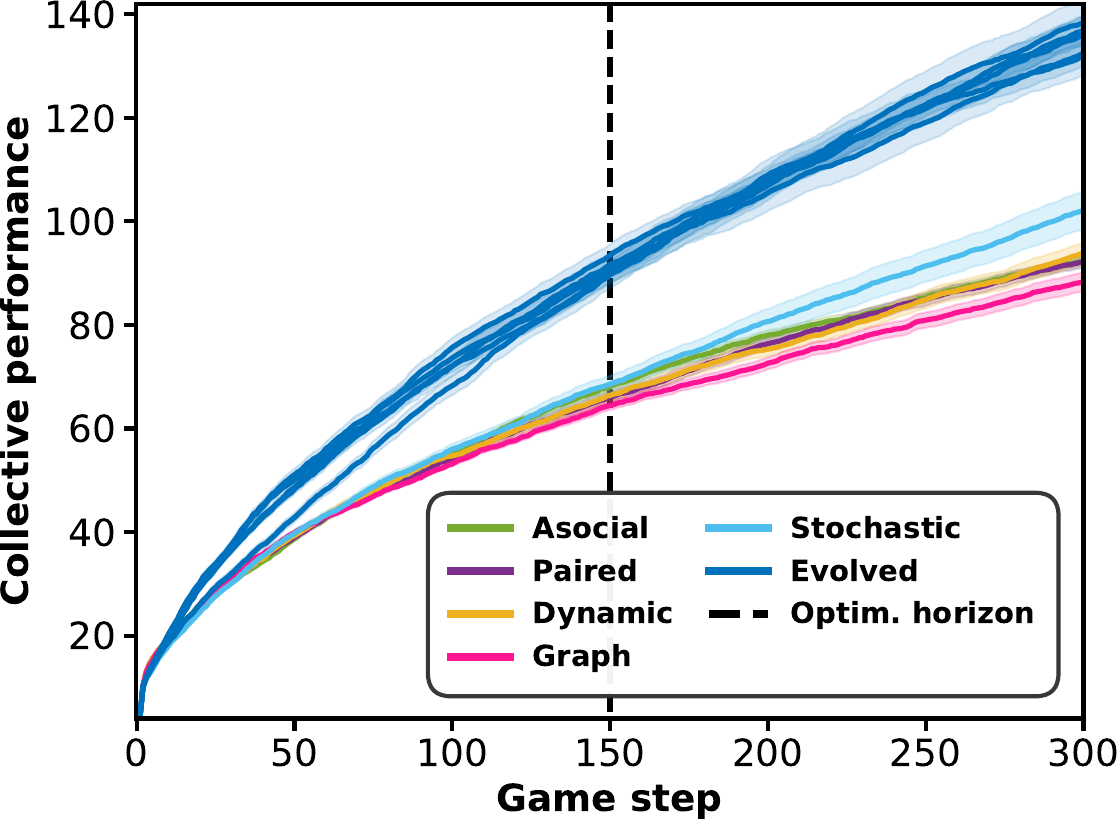}
        \caption{\label{fig:fitness_time}}
    \end{subfigure}\hfill
    \begin{subfigure}{0.49\linewidth}
        \centering
        \includegraphics[width=\linewidth]{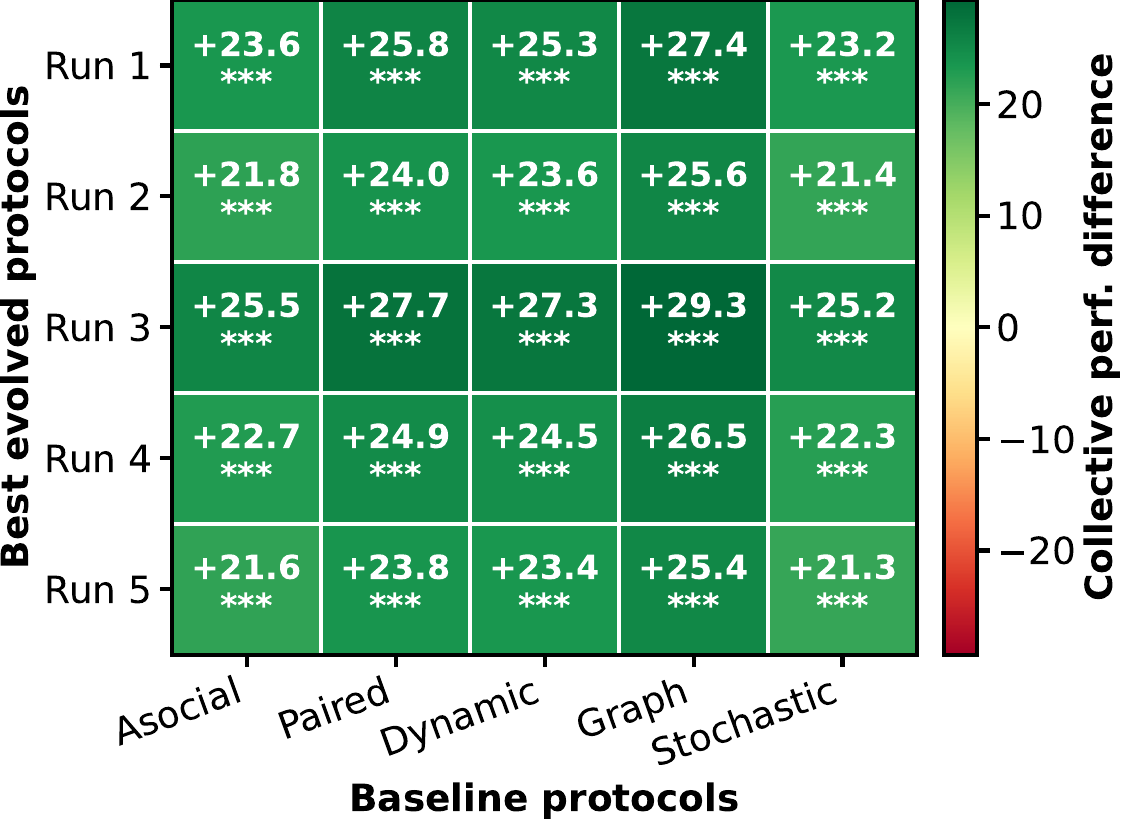}
        \caption{\label{fig:matrix}}
    \end{subfigure}
    \caption{\textbf{LLM-guided search uncovers high-performing transmission protocols}. (a) Collective performance of best protocols (one per evolutionary run) and baselines across game steps (mean$\pm$sem over 30 simulations). (b) Comparison of performance measures at 150 game steps. Stars indicate significance of one-sided t-tests (all $p<0.001$). \label{fig:optimization}}
\end{figure}

\subsection{LLM-guided search discovers high-performing transmission protocols}
\label{sec:res_performance}
We run 5 independent evolutionary searches, each producing a best protocol. All five evolved protocols significantly outperform all baselines, with an average advantage of +33\% over the best baseline at step 150 (+37\% for the top protocol; Figure~\ref{fig:matrix}). This advantage holds beyond the optimization horizon, reaching +32\% on average at step 300 (+36\% for the top, Figure~\ref{fig:fitness_time}). Appendix~\ref{app:extended_baselines} demonstrates similar conclusions for an extended set of 10 network-based baseline protocols.

An embedding analysis of protocol source codes confirms that different evolutionary runs explore distinct regions of the design space, with high-performing protocols spread across these regions rather than clustered (Appendix~\ref{app:embeddings}).

\subsection{State-awareness drives protocol performance}
\label{sec:res_analysis}

\begin{figure}[hb]
    \centering
    \includegraphics[width=0.5\linewidth]{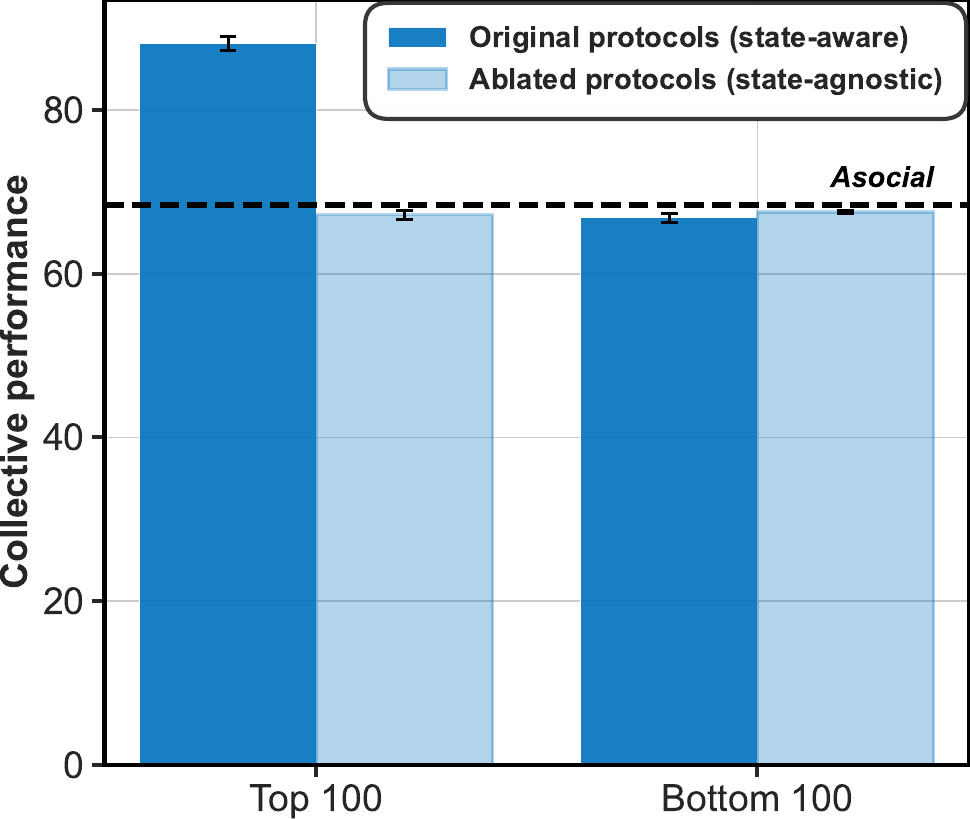}
    \caption{\textbf{State-awareness ablation}. Collective performance (mean$\pm$sem) for top-20 and bottom-20 protocols per evolutionary run (100 in each set), under original (full) and ablated (random content) conditions. Ablation eliminates the performance advantage of top protocols, reducing them to baseline levels, while bottom protocols are unaffected. \label{fig:ablation}}
\end{figure}

\paragraph{Performance requires state-awareness.} We ablate state-dependence by replaying each protocol's exact transmission schedule but randomly selecting which memories to send (see \textit{Methods}). We select the top and bottom 20 protocols for each of the five evolutionary runs (100 protocols in each set). For top protocols, ablation drops the average collective performance from 88 to 67 ($\Delta =-21$); performance on par with the asocial baseline (Figure~\ref{fig:ablation}). For bottom protocols, ablation has negligible effect ($\Delta=0.7$). High-performing protocols thus depend critically on state-aware content routing; the network topology and timing they produce provide no benefit on their own. Performance is also independent of transmission volume: fitness does not correlate with memories received per agent (Appendix~\ref{app:volume}).

\paragraph{How does state-awareness manifest?} Qualitative inspection of the best evolved protocols reveals state-awareness across transmission dimensions; full programs in Appendix~\ref{app:evolved_protocols}. \emph{What}: protocols filter by rarity, sharing only items owned by fewer than 20--40\% of agents (seeds 1, 2, 4), and prioritize items that appear frequently as ingredients in successful recipes (seeds 3, 5). Protocols also strategically share failures: seed 1 routes failed combinations to agents who possess both ingredients, preventing them from wasting a turn on the same attempt. Several protocols track sharing history to avoid re-sending recipes a recipient has already seen (seeds 1, 3, 5). \emph{Whom}: protocols route based on recipient state, \eg prioritizing recipes where the recipient owns one ingredient but not the result (seeds 3, 5). Multiple protocols partition agents into specialized groups (seeds 2, 3, 5), with group members sharing more intensively among themselves to deepen distinct branches of the discovery tree. \emph{When}: protocols adapt over time, narrowing the recipient pool as common items saturate the population (seeds 3, 4). Some shift from within-group to between-group sharing as the game progresses, initially building specialized knowledge and later cross-pollinating across groups (seeds 3, 5).

We validate these patterns quantitatively using the ablation-based correlational analysis described in \textit{Methods}. For 21 of 22 features, state-aware transmission contributes to fitness through measurable behavioral shifts (Table~\ref{tab:delta_results}). The strongest channel is recency bias ($\rho = 0.64$): protocols that use state-awareness to share recently discovered recipes achieve higher fitness. Recency interacts with recipient state ($\rho = 0.63$ when the recipient possesses both ingredients), suggesting successful protocols prioritize actionable recipes. Ingredient-based routing also contributes to fitness ($\rho = 0.40$--$0.43$), with stronger effects when shared items are novel to the recipient ($\rho$ up to $0.49$). Additional gains come from sharing rarer items ($\rho = 0.29$) and broadcasting widely rather than personalizing to individuals ($\rho = -0.28$). Effective protocols also adapt over time, increasing recency bias ($\rho = 0.45$) and ingredient targeting ($\rho = 0.44$) as the game progresses. Finally, state-aware routing shapes population-level diversity: effective protocols reduce early-game diversity ($\rho = -0.10$) but increase late-game diversity ($\rho = +0.12$), consistent with an isolate-then-share dynamic. Full results in Appendix~\ref{app:behavior_analysis}.


\subsection{Generalization across populations and domains}
\label{sec:res_generalization}

\begin{figure}[ht]
    \centering
    \includegraphics[width=\linewidth]{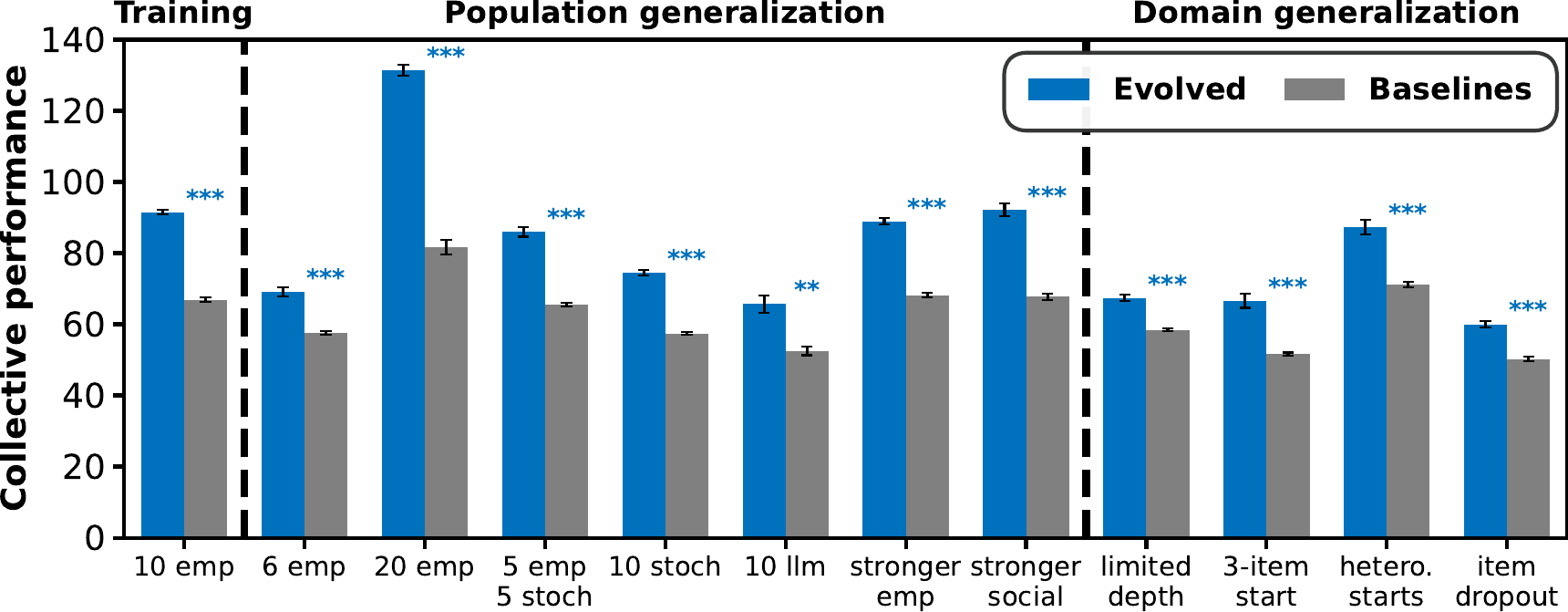}
    \caption{\textbf{Generalization across population and domain configurations}. Evolved protocols (blue) outperform baseline protocols (gray) across all population configurations and domain variants. Mean$\pm$sem, N=5, p<0.01 (**), p<0.001 (***).}
    \label{fig:generalization}
\end{figure}

We evaluate the best evolved protocol from each of the 5 runs on configurations they were not optimized for (Figure~\ref{fig:generalization}).

\paragraph{Population generalization.} Evolved protocols consistently outperform baselines across all population configurations, despite being optimized exclusively on 10 empowerment agents. Protocols transfer to smaller (6 agents: $+17\%$) and larger populations (20 agents: $+49\%$), to different agent types (stochastic: $+18\%$; LLM: $+16\%$; mixed: $+24\%$), and to agents with modified behavioral parameters (stronger empowerment bias: $+25\%$; stronger social bias: $+27\%$).

\paragraph{Domain generalization.} Protocols also generalize across modified versions of the discovery task, maintaining their advantage in all variants (depth-limited: $+12\%$; 3-item starts: $+25\%$; heterogeneous starts: $+20\%$; item dropout: $+18\%$).

\section{Discussion}
\label{sec:discussion}

\paragraph{Discovered protocols rely on state-awareness.}
Ablations confirm that state-aware content routing, not network topology or timing, drives protocol effectiveness. The same protocols transfer across agent types and domain variants, suggesting they exploit structural properties of the collective search problem rather than agent-specific idiosyncrasies.

\paragraph{Better understanding transmission.}
This work opens questions that could not be addressed within the network paradigm. Which dimensions of state-awareness matter most? How does state-awareness relate to mechanisms like transient diversity~\citep{smaldino2024maintaining} or synergistic information processing~\citep{garg2025causal}? Our ablation-based correlational analysis offers initial evidence: recency bias, ingredient-based routing, and temporal adaptation are the strongest state-dependent predictors of fitness, while knowledge diversity follows an isolate-then-share pattern consistent with transient diversity. The program search framework enables more targeted investigations, \eg ablating specific protocol features or constraining the design space to isolate what matters.

\paragraph{Limitations.}
Collective Little Alchemy, while exhibiting cumulative discovery and path dependencies, is a simplified domain with deterministic rules. Whether discovered protocols transfer to more complex or open-ended domains remains to be tested. Discovered protocols also assume centralized access to all agent states; whether effective protocols can be discovered under partial observability is an open question. Finally, we have not yet tested whether protocols discovered in silico transfer to human groups. 

\paragraph{Future directions.}
Deep mechanism design has demonstrated that interventions discovered in simulation can transfer to human participants~\citep{koster2022human}, suggesting this path is feasible for transmission protocols as well. Testing the transfer of our evolved protocols to human groups is a natural next step. Deployment could take the form of AI-mediated collaboration platforms that route information among participants, with humans retaining full autonomy over their decision-making. Anonymization could mitigate privacy concerns by routing based on what participants know rather than who they are.

Beyond transmission, program search could be applied to richer coordination design spaces: discovering organizational structures for crowdsourcing complex work~\citep{valentine2017flash} or effective mediation strategies for synthesizing group opinions~\citep{tessler2024ai}. More broadly, this work illustrates how AI systems can enhance human collective intelligence not by replacing human judgment, but by shaping the processes through which groups innovate~\citep{brinkmann2023machine, cui2024ai}. Program search with simulation scoring offers a general method for discovering such interventions, turning AI into infrastructure for human coordination rather than a substitute for it.

\subsection*{Acknowledgements}This research is partly funded by European Union’s Horizon 2020 research and innovation programme under the Marie Skłodowska-Curie grant agreement No 101065949 (CC).

\printbibliography

\clearpage
\appendix
\appendixpage

\setcounter{secnumdepth}{3} 
\setcounter{section}{0}
\renewcommand\thesection{\Alph{section}}
\renewcommand\thesubsection{\thesection.\arabic{subsection}}
\renewcommand\thesubsubsection{\thesubsection.\arabic{subsubsection}}
\renewcommand\sectionautorefname{Appendix}
\renewcommand\subsectionautorefname{Appendix}


\section{Additional Methods}
\label{app:additional_methods}

\subsection{Empowerment agent details}
\label{app:agents}

At each game step $t$, agents select a combination $(e_1, e_2)$ via softmax over empowerment values:
$$P(\text{choose } (e_1, e_2) \mid \text{agent } i, t) \propto \exp\left( \text{Emp}_i(e_1, e_2, t) / \beta \right),$$

with temperature $\beta = 0.05$. Combinations with zero empowerment are excluded from the candidate set.

\paragraph{Empowerment of attempting a combination.} The empowerment of attempting a combination is the expected empowerment of its outcome:
\begin{align*}
\text{Emp}_i(e_1, e_2, t) &= P(\text{success} \mid e_1, e_2, t) \\
&\quad \cdot \sum_{e_r} P(e_r \mid e_1, e_2, t) \cdot \text{Emp}_i(e_r, t).
\end{align*}
This requires estimates of: 1) the success probability $P(\text{success} \mid e_1, e_2, t)$, 2) the transition distribution $P(e_r \mid e_1, e_2, t)$ and 3) the empowerment of obtaining element $e_r$.

\paragraph{Empowerment of obtaining an element.} The empowerment of obtaining element $e$ for agent $i$ with its inventory $inv_i$ at game step $t$ is the number of unique elements that can be produced using $e$ as an ingredient, or zero if the agent already owns $e$:
$$\text{Emp}_i(e, t) = 
\begin{cases} 
0 & \text{if } e \in \text{inv}_i(t), \\ 
|\{e'' : \exists e',\, (e, e') \rightarrow e''\}| & \text{otherwise.} 
\end{cases}$$
To create heterogeneity across agents, we add multiplicative Gaussian noise ($\mu=1, \sigma=0.1$) to each agent's initial element empowerment estimates.

\paragraph{Success probability and transition distribution.} Both are estimated following the procedure of Brändle et al.~\citep{brandle_empowerment_2023}. They train neural networks using FastText word embeddings of element names to predict combination outcomes. Using 10-fold cross-validation, they train on 90\% of all possible transitions and predict outcomes for the remaining 10\% (success and distribution of resulting element), repeating this across all folds and aggregating predictions to obtain estimates of success probability and transition distribution for every possible combination.

\paragraph{Updating estimates during play.} As agents accumulate knowledge from their own attempts and social memories, estimates are updated as follows:
\begin{itemize}
    \item \textit{Failed combinations} (own or social): the combination becomes a deterministic failure, $P(\text{success} \mid e_1, e_2) = 0$, so $\text{Emp}_i(e_1, e_2, t) = 0$.
    \item \textit{Successful own attempts}: the outcome is now deterministic and the agent owns the result, so $\text{Emp}_i(e_r, t) = 0$.
    \item \textit{Successful social memories}: the outcome is now deterministic with $\text{Emp}_i(e_r, t)$ equal to the empowerment of the known result element.
\end{itemize}

\paragraph{Social bias.} When an agent's social inventory is non-empty, there is probability $p_{\text{social}}=0.5$ that it restricts its candidate set to combinations involving at least one socially received element.

\paragraph{Attention bottleneck.} Each agent receives at most 20 memories per step, subsampled uniformly when more are sent. This models bounded attention and ensures that protocol effectiveness depends on routing information strategically rather than broadcasting everything.

\subsection{LLM-guided evolutionary search details}
\label{app:evolution}

\paragraph{Archive structure.} The algorithm maintains an archive of protocols, each represented as executable Python code together with its fitness score and a natural-language description. The archive is seeded with the five baseline protocols and grows as new valid protocols are added. All valid protocols enter the archive regardless of fitness, preserving diversity for future exploration.

\paragraph{Warmup phase.} The first 50 iterations use a \emph{generation} mode that seeds the archive with diverse starting points. Each iteration, the LLM receives the five baseline protocols (code, descriptions, and fitness) as examples and is prompted to design a novel transmission protocol. Each valid protocol is added to the archive, bootstrapping diversity before refinement begins.

\paragraph{Evolution phase.} After warmup, each iteration selects a parent protocol from the archive, samples contrast solutions for inspiration, and prompts the LLM to design a new protocol.

\textit{Parent selection} uses Thompson sampling over fitness-based beliefs~\citep{tang2024code}. Each protocol's fitness is normalized to $[0,1]$ based on the current archive's minimum and maximum. A selection score is then sampled from a Beta distribution:
\begin{equation*}
    s \sim \text{Beta}(1 + \tilde{f} \cdot \lambda,\; 1 + (1-\tilde{f}) \cdot \lambda),
\end{equation*}
where $\tilde{f}$ is the normalized fitness and $\lambda=3$ controls selection pressure. The protocol with the highest sampled score is selected as parent. This balances exploitation of high-fitness solutions with exploration: even lower-fitness protocols retain a chance of selection.

\textit{Contrast solutions.} To expose the LLM to diverse strategies beyond the parent's lineage, we sample two \emph{contrast solutions} from the archive. Candidates exclude the parent and its ancestors, preventing redundant information. Selection uses a softmax over fitness-based ranks with temperature $\tau=0.4$, maintaining a soft preference for higher-performing solutions while preserving diversity.

\textit{Mutation modes.} We use two mutation modes with equal probability. In \emph{improve} mode, the LLM receives the parent protocol with full details\,---\,code, description, and fitness\,---\,framed as a solution to refine, with contrast solutions presented as alternative approaches for inspiration. In \emph{diversify} mode, all three protocols are presented equally (code, description, and fitness) as ``approaches that have been explored,'' and the LLM is prompted to propose a fundamentally different strategy. This encourages genuine exploration rather than incremental modification anchored to a single parent. Full prompts in Appendix~\ref{app:prompts}.

\paragraph{Validation.} Each candidate protocol undergoes validation before entering the archive. The LLM's response is parsed to extract the Python class, which is executed in a sandboxed simulation with a 20-second timeout. If execution fails or times out, the LLM is prompted to repair the code (up to three attempts). Valid protocols are evaluated for fitness: mean collective performance across 30 simulations with a population of 10 empowerment agents playing for 150 game steps.

\subsection{State-awareness ablation details}
\label{app:ablation}

To test whether protocol performance stems from state-aware routing decisions or from structural properties of induced transmission patterns, we construct a state-agnostic ablation of each protocol. The ablation replays the exact transmission schedule of the original protocol\,---\,who sends to whom, how many memories, and at which step\,---\,but randomly selects which memories to send from the sender's available memories. Because agent states diverge across runs (different content leads to different discoveries and inventories), the replayed schedule is effectively state-agnostic: the original protocol's decisions about what to share, with whom, and when may have been functions of agent and collective states, but the ablated schedule is fixed regardless of what agents actually know or discover.

\paragraph{Protocol selection.} We select the top and bottom 20 protocols from each of 5 generation runs, ranked by mean fitness on a held-out subset of evaluation seeds (evaluation seeds 0--14).

\paragraph{Matched comparisons.} For each selected protocol, we run matched comparisons on the remaining 15 evaluation seeds (seeds 15--29). Each seed produces two trajectories: one using the original protocol and one using the schedule-matched but state-agnostic ablation. This paired design allows direct comparison of fitness while controlling for stochastic variation across seeds.

\paragraph{Shadow log analysis.} To confirm that observed behavioral signatures stem from state-aware programming rather than schedule structure, we compute the same behavioral features (defined in Appendix~\ref{app:features}) on both the original and ablated simulation logs. We compare effect sizes (Cohen's $d$) between original and ablated feature distributions, and compare fitness correlations (Spearman rank) across the two conditions. Features that show large effect sizes (Cohen's $d > 2.0$) and substantial drops in fitness correlation ($\Delta\rho$) between original and ablated logs are interpreted as signatures of state-aware programming.

\subsection{Behavioral feature definitions}
\label{app:features}
Table~\ref{tab:features} lists the 22 behavioral features used in our analysis. For each transmission event $(t, \text{teacher}, \text{learner}, \text{recipe})$, we compute per-transmission features and aggregate them (mean) across all transmissions and evaluation seeds. Temporal window variants split the simulation into thirds (early, mid, late). The simulation is also run under content-randomized ablation (same schedule, random content); for each feature, we report the Spearman correlation between the per-protocol original--ablated difference and fitness (Table~\ref{tab:delta_results}).

\begin{table*}[ht]
\centering
\small
\caption{Behavioral features computed from transmission logs. Each feature is averaged across all transmissions (or per-step aggregations) and then across evaluation seeds. Notation: recipe $= (i_1, i_2, r)$ where $i_1, i_2$ are ingredients and $r$ is the result ($r = \text{null}$ for failures). Features indexed by $k \in \{0,1,2\}$ are computed separately for each value.}
\label{tab:features}
\begin{tabular}{llp{8cm}}
\toprule
\textbf{Group} & \textbf{Feature} & \textbf{Definition} \\
\midrule
\multirow{5}{*}{\rotatebox{90}{Content}}
& Recency bias & Discovery step of shared recipe / transmission step. Higher values indicate sharing recent discoveries. \\
& Rarity bias & Among successful recipes, $1 - (\text{fraction of agents who own } r)$. Higher values indicate sharing rarer items. \\
& Item novelty & Among successful recipes, fraction where $r$ is not in the learner's inventory. \\
& Actionable has-$k$ & Fraction of transmissions where the learner owns exactly $k$ ingredients AND the result item is novel to the learner. Combines routing with content novelty. \\
& Success bias & Whether the shared recipe is successful ($r \neq \text{null}$) minus the teacher's base success rate at that step. \\
\midrule
\multirow{5}{*}{\rotatebox{90}{Routing}}
& Has-$k$ fraction & Fraction of transmissions where the learner owns exactly $k$ ingredients of the shared recipe. \\
& Cond.\ recency has-$k$ & Recency bias computed separately for transmissions where the learner owns exactly $k$ ingredients. Captures the interaction between content recency and recipient state. \\
& Targeted failure rate & Among shared failures, fraction where the learner owns both ingredients (can verify the failure). \\
& Personalization & Mean Jaccard distance between recipe sets sent by the same teacher to different learners at the same step. Higher values indicate more tailored transmissions. \\
& Exclusive routing & Mean $1 / n_{\text{recipients}}$ for each (recipe, step). Higher values indicate recipes are sent to fewer learners. \\
\midrule
\multirow{1}{*}{\rotatebox{90}{\scriptsize Syn.}}
& Knowledge synergy & Population-level mean pairwise Jaccard distance across all agent pairs at window boundaries (excluding initial elements). Computed for early and late windows. \\
\midrule
\multirow{3}{*}{\rotatebox{90}{Temporal}}
& Discovery burst rate & Mean outgoing transmissions on steps where the teacher made a discovery / mean on non-discovery steps. Values ${>}1$ indicate discovery-triggered sharing. \\
& Slope recency & Late-window recency bias minus early-window recency bias. Positive values indicate increasing recency preference over time. \\
& Slope has-$k$ & Late-window has-$k$ fraction minus early-window has-$k$ fraction. Captures temporal shifts in ingredient-based routing. \\
\bottomrule
\end{tabular}
\end{table*}


\label{app:additional_results}

\section{Additional Results}

\subsection{Protocol fitness across evolution}
\begin{figure}[H]
    \centering
    \includegraphics[width=0.8\linewidth]{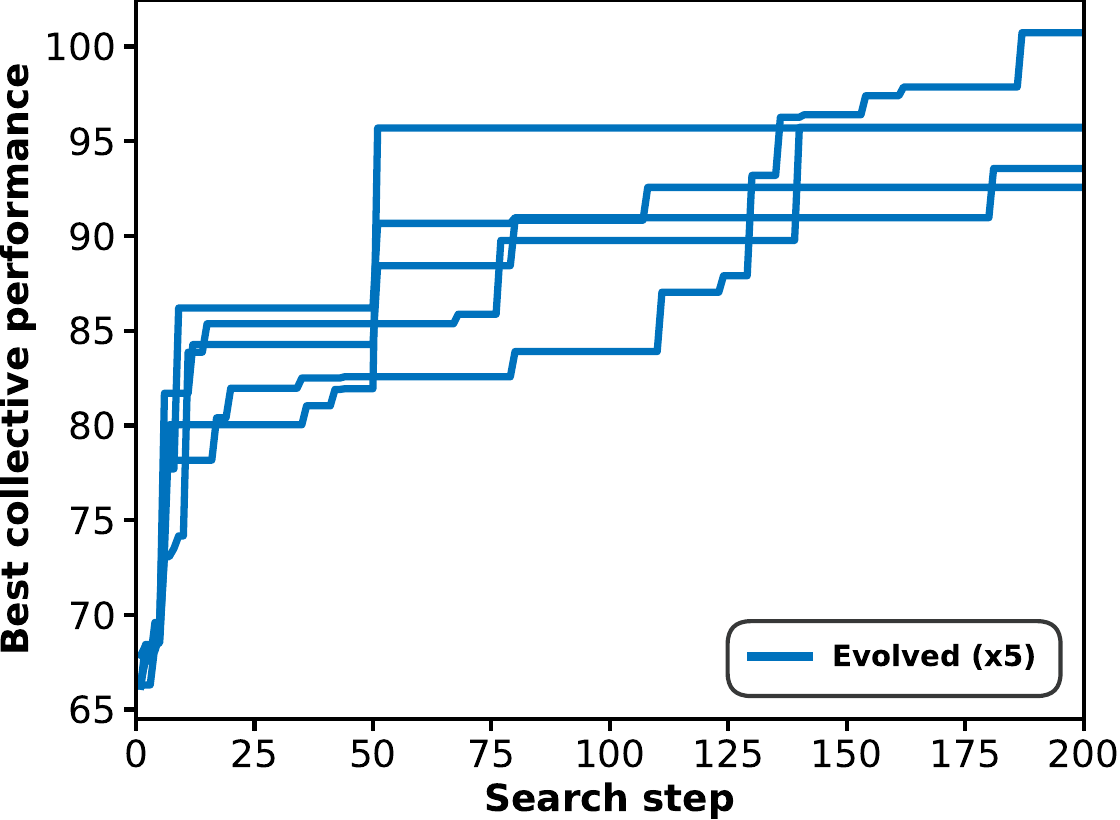}
    \caption{\textbf{Fitness of best evolved protocol as a function of search step}.}
    \label{fig:app_max_fitness}
\end{figure}

Figure~\ref{fig:app_max_fitness} shows the fitness of the best-performing protocol as a function of search iterations, for each of 5 generation runs and 5 evolution runs. Starting from the baseline protocols used to seed the archive (steps 1--5), both search processes progressively discover better protocols, with fitness increasing from 66--70 elements discovered after 150 game steps (baselines) to 85--97 elements ($+21$--$38\%$ over best baseline). 

\subsection{Extended baseline experiments}
\label{app:extended_baselines}
 We evaluate protocols against an extended set of baselines drawn from the cultural evolution and collective intelligence literature:
\begin{itemize}
    \item \textit{Asocial}: agents explore independently with no information sharing, providing a lower bound.
    \item \textit{Stochastic}: each agent receives the maximum number of memories selected at random from randomly selected peers, representing maximal undirected sharing ($n_\text{mem}=20$ per teacher, 20 random draws per step).
    \item \textit{Paired}: agents are assigned to fixed dyads and exchange memories only with their partner ($n_\text{mem}=10$, $p_\text{share}=0.5$), mirroring simple reciprocal teaching~\citep{nisioti_social_2022, nisioti2024collective}.
    \item \textit{Dynamic}: agents maintain fixed pairs but periodically visit other dyads ($n_\text{mem}=10$, $p_\text{share}=0.5$), capturing fission--fusion social dynamics~\citep{nisioti_social_2022, nisioti2024collective}.
    \item \textit{Random graph}: agents communicate along edges of an Erd\H{o}s--R\'enyi random graph ($p_\text{conn}=0.2$, $n_\text{mem}=10$, $p_\text{share}=0.5$)~\citep{lazer2007network, mason2008propagation}.
    \item \textit{Ring}: agents are arranged on a ring lattice connected to their $k=4$ nearest neighbors (Watts--Strogatz with $p=0$; $n_\text{mem}=8$, $p_\text{share}=0.7$), a topology widely studied in models of cumulative cultural evolution~\citep{cantor2021social, moser2023innovation}.
    \item \textit{Small-world}: a Watts--Strogatz graph ($k=4$, $p_\text{rewire}=0.3$; $n_\text{mem}=12$, $p_\text{share}=0.6$) that introduces shortcut edges to reduce path lengths while preserving local clustering~\citep{cantor2021social, moser2023innovation, garg2025causal}.
    \item \textit{Caveman}: a connected caveman graph of fully connected cliques (size 5) joined by single inter-clique bridges ($n_\text{mem}=15$, $p_\text{share}=0.4$), representing modular social organization~\citep{cantor2021social, moser2023innovation}.
    \item \textit{Multilevel}: a hierarchical network with two super-groups each containing two sub-modules, with edge probabilities decreasing across levels ($p_\text{within}=1.0$, $p_\text{between-modules}=0.3$, $p_\text{between-supergroups}=0.1$; $n_\text{mem}=6$, $p_\text{share}=0.8$), inspired by the nested band--camp structure of hunter-gatherer societies~\citep{migliano2020hunter, cantor2021social}.
    \item \textit{Intermittent}: agents alternate between three rounds of independent exploration and one round of full broadcast ($n_\text{mem}=2$ per teacher, $p_\text{share}=1.0$), following evidence that intermittent breaks in interaction improve collective intelligence~\citep{bernstein2018intermittent}.
\end{itemize}

\begin{figure}[h!]
    \centering
    \includegraphics[width=0.8\columnwidth]{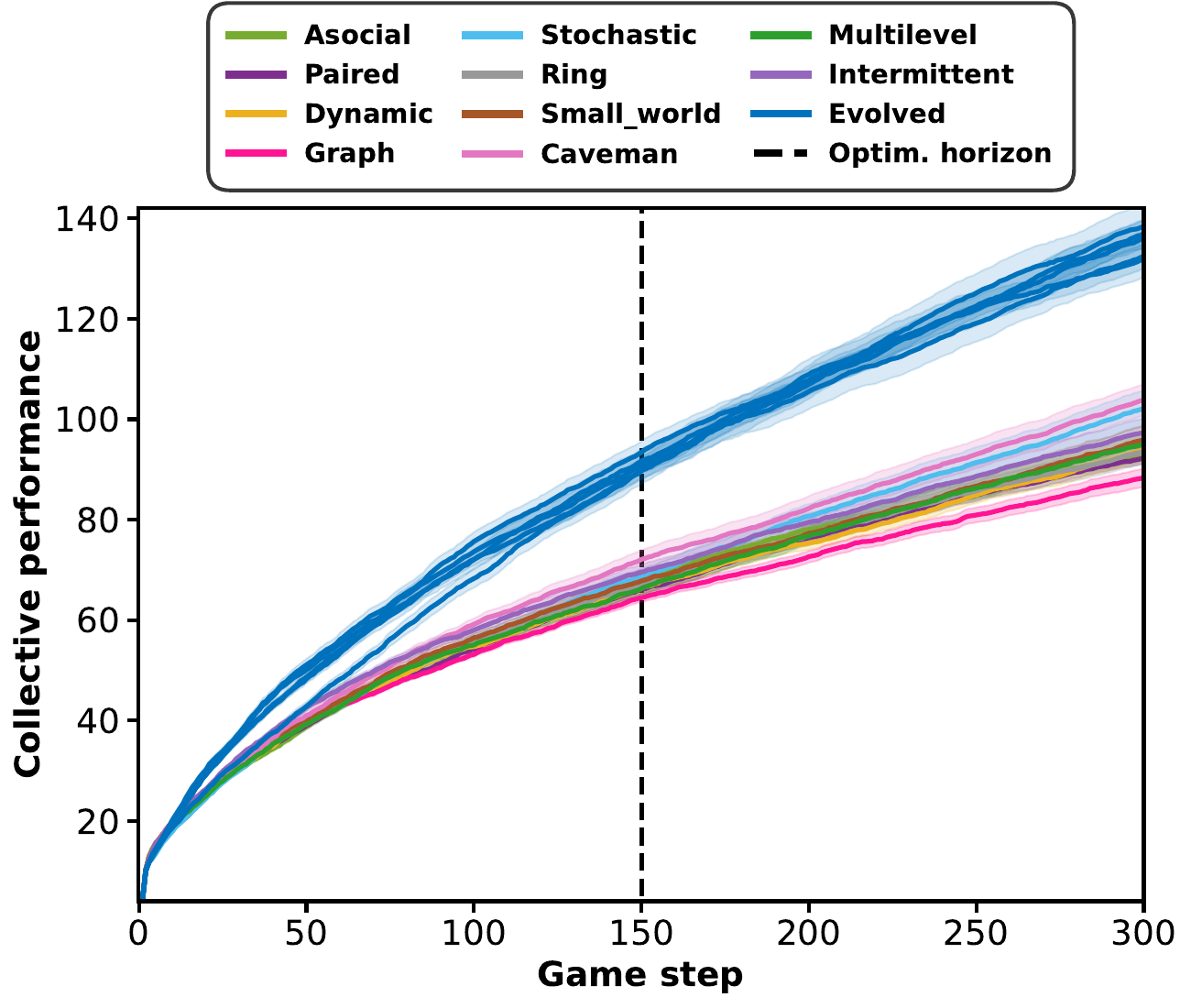}
    \caption{\textbf{LLM-guided search uncovers high-performing protocols for populations of 10 empowerment agents}. Fitness of best protocols (one per search run) and extended set of baselines across 300 game steps (mean$\pm$sem over 30 simulations).}
    \label{fig:extended_baselines}
\end{figure}

All social baselines use random content selection (no state-awareness), isolating the effect of network topology from content-selection strategy. Protocol programs can be found in \texttt{src/domains/transmission\_protocols.py} in the codebase. Figure~\ref{fig:extended_baselines} shows that none of these baselines reaches performance levels comparable to the set of best evolved and generated protocols (one per search run). 

\subsection{Evolution explores beyond the generative prior.}
\label{app:embeddings}

We examine how evolved and generated protocols distribute across the design space using high-dimensional embeddings of protocol source code (OpenAI \texttt{text-embedding-3-large} model). We visualize t-SNE projections~\citep{maaten2008visualizing} and measure distributional divergence using maximum mean discrepancy (MMD) with a cosine similarity kernel~\citep{gretton2012kernel}. 

\begin{figure}[h!]
    \centering
    \includegraphics[width=0.99\columnwidth]{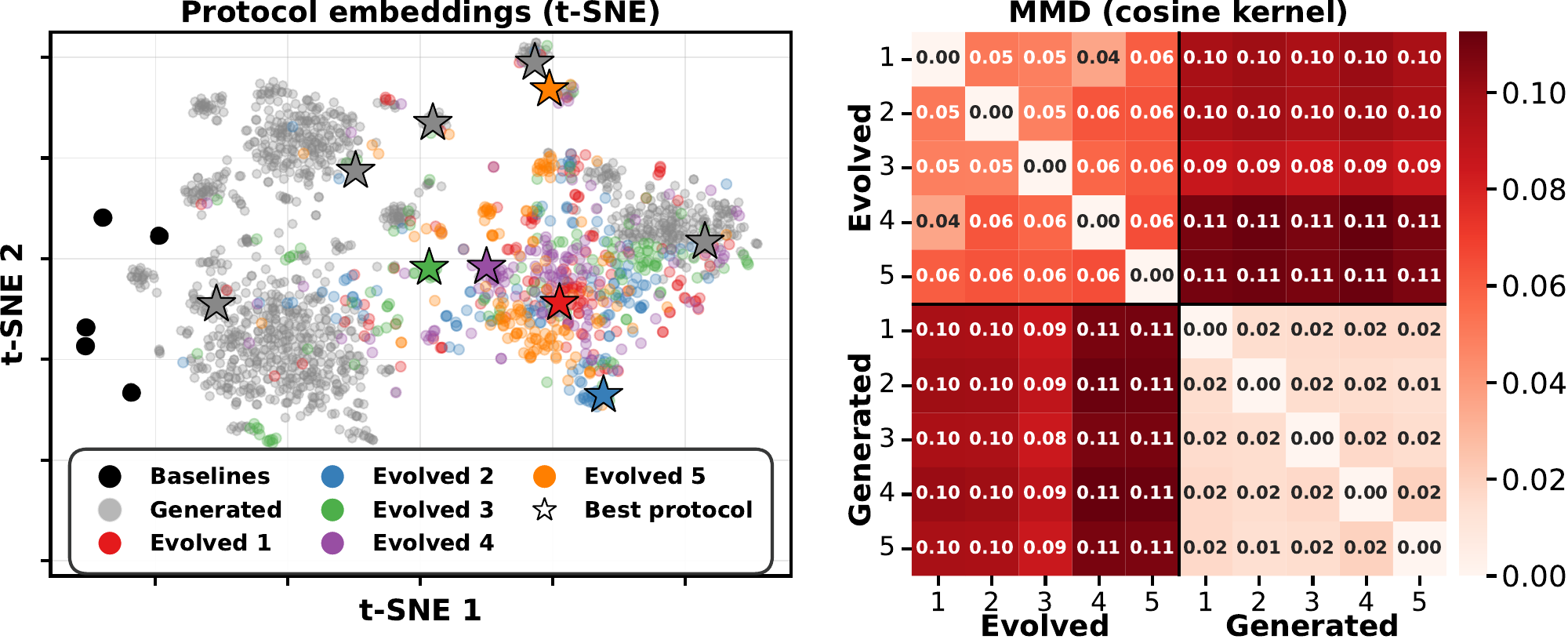}
    \caption{\textbf{Program search discovers transmission protocols that diverge from the generative prior}. (left) t-SNE projections of discovered protocols, with per-run best-performing protocols marked with stars. (right) Maximum mean discrepancy (MMD) measuring distributional divergence across search runs; evolved protocols are more distant from generated protocols than from each other.}  
    \label{fig:embedding}
\end{figure}

Both generated protocols and the warmup phase of evolution sample from the same prior: an LLM conditioned on baseline protocols (grey points in Figure~\ref{fig:embedding}, left). After warmup, evolution diversifies and improves upon initially generated protocols, exploring regions beyond the LLM prior. The MMD matrix (Figure~\ref{fig:embedding}, right) quantifies this: divergence between generated seeds is low, as expected from the shared prior. Evolved--generated divergence is substantially higher, confirming that evolution pushes protocols away from the prior. Evolved--evolved divergence also exceeds generated--generated divergence, indicating that different seeds explore distinct regions. The best-performing protocols for each run (stars) are spread across these regions rather than clustered, suggesting multiple distinct strategies can achieve high performance.

\subsection{Transmission effectiveness is not correlated with volume}
\label{app:volume}

\begin{figure}[h]
    \centering
    \includegraphics[width=0.5\columnwidth]{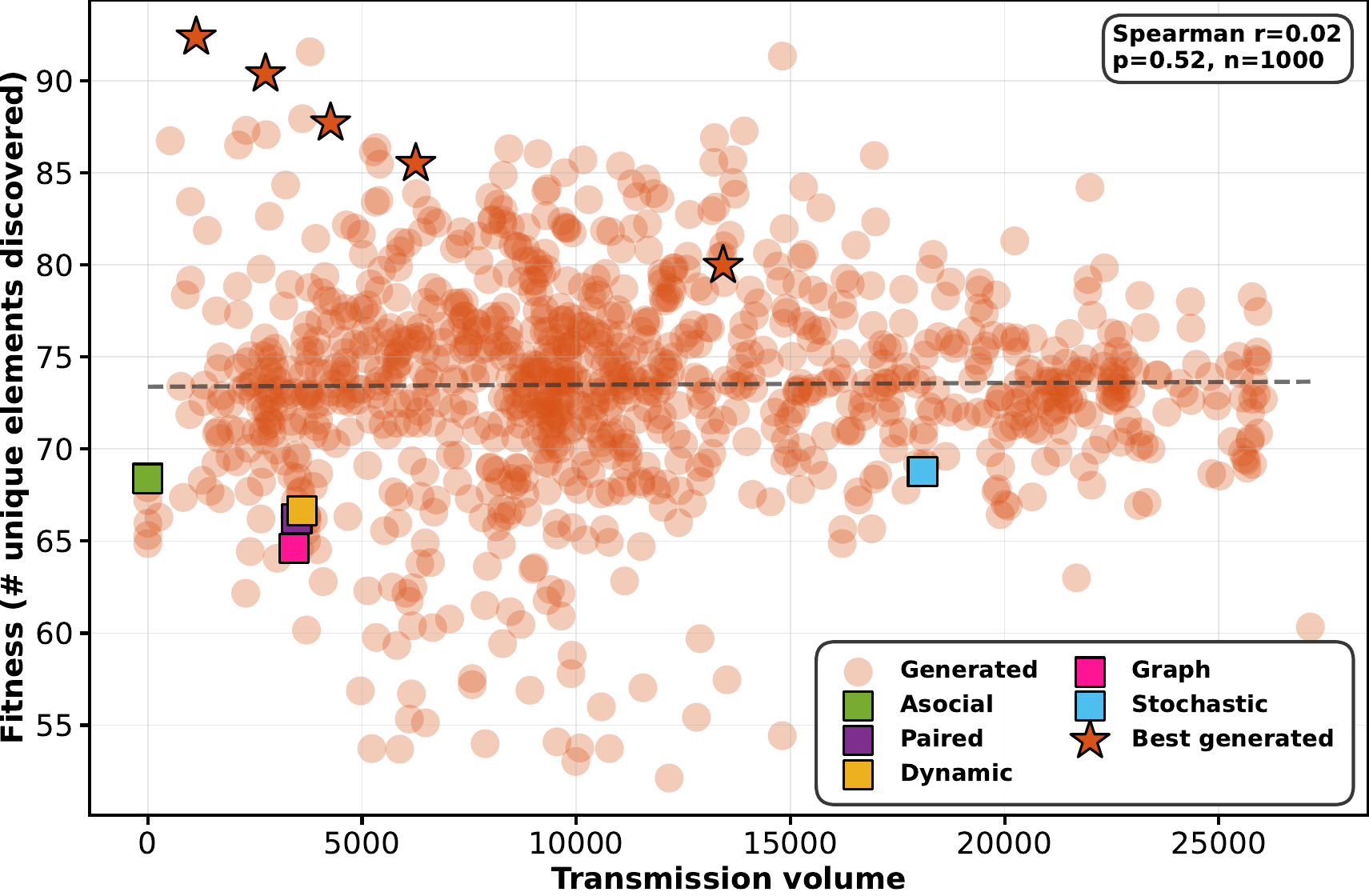}
    \caption{\textbf{Transmission protocol effectiveness is not correlated with transmission volume}. The quantity of information transmitted is measured as the number of memories received by an agent that had not experienced this memory on its own at the time of reception.}
    \label{fig:memory_scatter}
\end{figure}

Figure~\ref{fig:memory_scatter} shows no correlation between fitness and the number of novel memories received per agent across 997 generated protocols, consistent with prior findings that more sharing does not necessarily improve collective performance~\citep{bernstein2018intermittent, derex_partial_2016, nisioti_social_2022}. 

\clearpage
\subsection{Behavioral feature analyses.}
\label{app:behavior_analysis}
Table~\ref{tab:delta_results} presents all results for our behavioral analysis. 

\begin{table}[ht]
    \centering
    \small
    \caption{Ablation analysis results across ${\sim}1{,}000$ independently generated protocols. For each feature, we compute the per-protocol difference between original and content-randomized (ablated) values, and correlate this difference with fitness (Spearman $\rho$). $\boldsymbol{n}$ indicates the number of protocols for which the feature was defined. Significance after Benjamini--Hochberg FDR correction: ** $p < 0.01$, *** $p < 0.001$. Features sorted by $|\rho|$.}
    \label{tab:delta_results}
    \begin{tabular}{llrrrl}
        \textbf{Group} & \textbf{Feature} & $\boldsymbol{\rho}$ & \textbf{Sig.} \\
        \midrule
        Content & Recency bias             & $+0.64$ & *** \\
        Routing & Cond.\ recency has-2     & $+0.63$ & *** \\
        Routing & Cond.\ recency has-1     & $+0.58$ & *** \\
        Routing & Targeted failure rate    & $-0.54$ & *** \\
        Content & Actionable has-0         & $+0.49$ & *** \\
        Routing & Cond.\ recency has-0     & $+0.48$ & *** \\
        Temporal & Slope recency           & $+0.45$ & *** \\
        Temporal & Slope has-1             & $+0.44$ & *** \\
        Routing & Has-2 fraction           & $-0.43$ & *** \\
        Routing & Has-1 fraction           & $+0.42$ & *** \\
        Content & Actionable has-1         & $+0.41$ & *** \\
        Routing & Has-0 fraction           & $+0.40$ & *** \\
        Temporal & Slope has-2             & $-0.39$ & *** \\
        Content & Rarity bias              & $+0.29$ & *** \\
        Routing & Personalization          & $-0.28$ & *** \\
        Routing & Exclusive routing        & $-0.26$ & *** \\
        Temporal & Slope has-0             & $+0.22$ & *** \\
        Content & Item novelty             & $+0.22$ & *** \\
        Content & Actionable has-2         & $+0.17$ & *** \\
        Synergy & Knowledge synergy (late) & $+0.12$ & *** \\
        Synergy & Knowledge synergy (early)& $-0.10$ &  ** \\
        Temporal & Discovery burst rate    & $+0.09$ &  ** \\
        \midrule
        Content & Success bias             & $+0.00$ &     \\
        \bottomrule
    \end{tabular}
\end{table}

\onecolumn
\section{Prompts and Example Protocols}
\label{app:prompts_and_protocols}
\subsection{Evolved protocols}
\label{app:evolved_protocols}
\begin{tcbverbatim}[Evolved protocol, seed 1]{basicstyle=\ttfamily\scriptsize}
import numpy as np
import random
from collections import defaultdict, Counter

class TransmissionProtocol:
    def __init__(self, n_agents, n_steps):
        self.n_agents = n_agents
        self.n_steps = n_steps
        self.history_shared = defaultdict(set)

    def share_memories(self, i_step, agent_states):
        memory_exchanges = {aid: [] for aid in range(self.n_agents)}
        
        # 1. Map current knowledge
        inventories = {aid: set(state['inventory']) for aid, state in agent_states.items()}
        global_items = set()
        for inv in inventories.values():
            global_items.update(inv)
        
        item_prevalence = Counter()
        for inv in inventories.values():
            item_prevalence.update(inv)

        # 2. Extract and categorize memories
        success_map = defaultdict(list) # item -> [(teacher, midx)]
        failure_map = [] # list of (teacher, midx, it1, it2)
        
        for aid, state in agent_states.items():
            for m_idx, (it1, it2, res) in enumerate(state['memories']):
                if res is not None:
                    if item_prevalence[res] < self.n_agents * 0.25:
                        success_map[res].append((aid, m_idx))
                else:
                    failure_map.append((aid, m_idx, it1, it2))

        # 3. Strategic Distribution
        # Sort items by rarity (rarest first)
        rare_items = sorted(success_map.keys(), key=lambda x: item_prevalence[x])
        
        # Grid-Partitioning: Divide agents into 4 "Logical Niches"
        # Each niche is encouraged to focus on different basic combos
        for learner_id in range(self.n_agents):
            learner_inv = inventories[learner_id]
            slots = []
            
            # Priority A: Success Relay
            # Distribute rare items such that different agents get different rare items
            # Use learner_id to offset the starting point in the rare_items list
            offset = learner_id % max(1, len(rare_items))
            for i in range(len(rare_items)):
                item_to_share = rare_items[(offset + i) % len(rare_items)]
                if item_to_share not in learner_inv and item_to_share not in self.history_shared[learner_id]:
                    teacher_info = random.choice(success_map[item_to_share])
                    slots.append(teacher_info)
                    self.history_shared[learner_id].add(item_to_share)
                    if len(slots) >= 12: # Leave room for failure signals
                        break
            
            # Priority B: Failure Signal (Ant-Path Pruning)
            # Only share a failure if the learner CAN actually perform that failure
            # This saves the learner from wasting a turn.
            random.shuffle(failure_map)
            for tid, mid, it1, it2 in failure_map:
                if len(slots) >= 20:
                    break
                if tid != learner_id:
                    if it1 in learner_inv and it2 in learner_inv:
                        slots.append((tid, mid))            
            memory_exchanges[learner_id] = slots
        return memory_exchanges

\end{tcbverbatim}
\begin{tcbverbatim}[Evolved protocol, seed 2]{basicstyle=\ttfamily\scriptsize}
import numpy as np
import random
from collections import defaultdict, Counter

class TransmissionProtocol:
    def __init__(self, n_agents, n_steps):
        self.n_agents = n_agents
        self.n_steps = n_steps
        self.basics = [0, 1, 2, 3] # Representing Air, Water, Earth, Fire
        self.agent_roles = {}
        self.silos = defaultdict(list)
        
        # Assign roles
        ids = list(range(n_agents))
        random.shuffle(ids)
        for i, aid in enumerate(ids):
            role = self.basics[i % 4]
            self.agent_roles[aid] = role
            self.silos[role].append(aid)

    def share_memories(self, i_step, agent_states):
        memory_exchanges = {aid: [] for aid in range(self.n_agents)}
        
        # Calculate global rarity for filtering
        item_occurrence = Counter()
        for state in agent_states.values():
            item_occurrence.update(state['inventory'])
        
        for learner_id in range(self.n_agents):
            learner_role = self.agent_roles[learner_id]
            learner_inv = set(agent_states[learner_id]['inventory'])
            
            candidates = []
            
            # 1. Intra-Silo Sharing (High Volume)
            silo_mates = self.silos[learner_role]
            for mate_id in silo_mates:
                if mate_id == learner_id: continue
                mems = agent_states[mate_id]['memories']
                # Take recent successes
                for mid in range(len(mems)-1, max(-1, len(mems)-15), -1):
                    m1, m2, res = mems[mid]
                    if res is not None:
                        if res not in learner_inv:
                            candidates.append((mate_id, mid, 2)) # Weight 2 for specialty
                            
            # 2. Inter-Silo Crossing (Selective)
            # Only take items that are rare or might be useful cross-specialty
            other_agents = random.sample(list(agent_states.keys()), min(10, self.n_agents))
            for teacher_id in other_agents:
                if self.agent_roles[teacher_id] == learner_role: continue
                
                mems = agent_states[teacher_id]['memories']
                # Take very few, high-value ones
                for mid in range(len(mems)-1, max(-1, len(mems)-5), -1):
                    m1, m2, res = mems[mid]
                    if res is not None and res not in learner_inv:
                        if item_occurrence[res] < (self.n_agents * 0.2): # Rare discovery
                            candidates.append((teacher_id, mid, 1)) # Weight 1
            
            # 3. Apply bottleneck with priority
            # Sort by weight (desc), then by recency (the mid is already high-to-low)
            candidates.sort(key=lambda x: x[2], reverse=True)
            
            selection = candidates[:20]
            for tid, mid, weight in selection:
                memory_exchanges[learner_id].append((tid, mid))
        return memory_exchanges

\end{tcbverbatim}
\begin{tcbverbatim}[Evolved protocol, seed 3]{basicstyle=\ttfamily\scriptsize}
import numpy as np
import random
from collections import defaultdict, Counter

class TransmissionProtocol:
    def __init__(self, n_agents, n_steps):
        self.n_agents = n_agents
        self.n_steps = n_steps
        # Assign agents to 4 lineages: 0: Air, 1: Water, 2: Earth, 3: Fire
        self.lineages = {aid: aid % 4 for aid in range(n_agents)}
        self.lineage_members = defaultdict(list)
        for aid, lin in self.lineages.items():
            self.lineage_members[lin].append(aid)
            
        self.history_sent = defaultdict(set) # (aid, result): bool
        self.usage_counts = Counter() # item: how many successful recipes it was a parent in

    def share_memories(self, i_step, agent_states):
        memory_exchanges = {aid: [] for aid in range(self.n_agents)}
        inv_sets = {aid: set(agent_states[aid]['inventory']) for aid in range(self.n_agents)}
        
        # Calculate pivotal items: items used as parents in teacher memories
        self.usage_counts.clear()
        for aid in range(self.n_agents):
            for m1, m2, res in agent_states[aid]['memories']:
                if res is not None:
                    self.usage_counts[m1] += 1
                    self.usage_counts[m2] += 1

        for lid in range(self.n_agents):
            learner_lin = self.lineages[lid]
            lin_peers = [p for p in self.lineage_members[learner_lin] if p != lid]
            other_peers = [p for p in range(self.n_agents) if self.lineages[p] != learner_lin]
            
            # Quota: Intra-lineage vs Inter-lineage
            is_early = i_step < self.n_steps * 0.6
            intra_quota = 16 if is_early else 8
            
            selected_mems = []
            
            # Helper to find items from a pool of teachers
            def collect_from(teachers, limit, priority_pivotal=False):
                pool = []
                for tid in teachers:
                    mems = agent_states[tid]['memories']
                    # Looking at recent discoveries
                    for mid in range(max(0, len(mems)-40), len(mems)):
                        m1, m2, res = mems[mid]
                        if res is None or res in inv_sets[lid]:
                            continue
                        if (lid, res) in self.history_sent:
                            continue
                        
                        score = 0
                        # Is it pivotal?
                        is_pivotal = self.usage_counts[res] > 1
                        if is_pivotal: score += 100
                        
                        # Can learner make it?
                        if m1 in inv_sets[lid] and m2 in inv_sets[lid]:
                            score += 50
                        elif m1 in inv_sets[lid] or m2 in inv_sets[lid]:
                            score += 10
                            
                        pool.append((tid, mid, res, is_pivotal, score + random.random()))
                
                pool.sort(key=lambda x: x[4], reverse=True)
                return pool[:limit]

            # 1. Intra-Lineage (The Base)
            lin_mems = collect_from(lin_peers, intra_quota)
            for m in lin_mems:
                selected_mems.append((m[0], m[1]))
                self.history_sent[(lid, m[2])].add(m[1])

            # 2. Inter-Lineage (The Cross-Pollination)
            inter_quota = 20 - len(selected_mems)
            other_mems = collect_from(other_peers, inter_quota, priority_pivotal=True)
            for m in other_mems:
                selected_mems.append((m[0], m[1]))
                self.history_sent[(lid, m[2])].add(m[1])
            memory_exchanges[lid] = selected_mems
        return memory_exchanges
\end{tcbverbatim}
\begin{tcbverbatim}[Evolved protocol, seed 4]{basicstyle=\ttfamily\scriptsize}
import numpy as np
import random
from collections import Counter, defaultdict

class TransmissionProtocol:
    def __init__(self, n_agents, n_steps):
        self.n_agents = n_agents
        self.n_steps = n_steps
        self.item_depths = {b: 0 for b in ['air', 'water', 'earth', 'fire']}
        self.global_counts = Counter()
        self.new_discoveries_prompted = 0

    def get_affinity(self, inventory):
        # Determine which basic element the agent uses most
        # Simplified: check for descendants or frequency
        return random.randint(0, 3) # Placeholder for more complex affinity logic

    def share_memories(self, i_step, agent_states):
        memory_exchanges = {aid: [] for aid in range(self.n_agents)}
        
        # 1. Update Global Census & Depths
        current_inv_sizes = []
        for aid, state in agent_states.items():
            inv = state['inventory']
            current_inv_sizes.append(len(inv))
            for item in inv:
                self.global_counts[item] += 1
            
            for m1, m2, res in state['memories']:
                if res is not None and res not in self.item_depths:
                    d1 = self.item_depths.get(m1, 0)
                    d2 = self.item_depths.get(m2, 0)
                    self.item_depths[res] = max(d1, d2) + 1

        # 2. Dynamic Thresholds
        progression = i_step / self.n_steps
        saturation_limit = 0.4 - (0.3 * progression) # Tightens over time
        
        # 3. Memory Filtering & Scoring
        for learner_id in range(self.n_agents):
            learner_inv = set(agent_states[learner_id]['inventory'])
            candidates = []
            
            # Potential Teachers (biased towards those with different but deep inventories)
            teachers = range(self.n_agents)
            
            for teacher_id in teachers:
                if teacher_id == learner_id: continue
                
                t_mems = agent_states[teacher_id]['memories']
                # Examine recent successes (frontier)
                for m_idx in range(max(0, len(t_mems)-30), len(t_mems)):
                    m1, m2, res = t_mems[m_idx]
                    
                    if res is None or res in learner_inv:
                        continue
                        
                    # Scoring Logic
                    depth = self.item_depths.get(res, 1)
                    rarity = 1.0 / (self.global_counts[res] + 1)
                    
                    is_actionable = 1 if (m1 in learner_inv and m2 in learner_inv) else 0
                    
                    # Boost score if it's actionable AND deep (The Frontier)
                    score = (depth ** 2) * rarity
                    if is_actionable:
                        score *= 10.0
                    else:
                        # If not actionable, only share if very deep (to prepare prerequisites)
                        score *= 0.5
                    
                    # Diversify slightly
                    score *= random.uniform(0.9, 1.1)
                    candidates.append((score, (teacher_id, m_idx)))

            # Sort and select top 20
            candidates.sort(key=lambda x: x[0], reverse=True)
            selection = [item[1] for item in candidates[:20]]
            memory_exchanges[learner_id] = selection
        return memory_exchanges
\end{tcbverbatim}
\begin{tcbverbatim}[Evolved protocol, seed 5]{basicstyle=\ttfamily\scriptsize}
import numpy as np
import random
from collections import defaultdict, Counter

class TransmissionProtocol:
    def __init__(self, n_agents, n_steps):
        self.n_agents = n_agents
        self.n_steps = n_steps
        self.n_guilds = 4
        self.agent_guilds = {aid: (aid % self.n_guilds) for aid in range(n_agents)}
        self.global_knowledge = {} # res -> (teacher_id, m_idx, (p1, p2))
        self.item_centrality = Counter() # item -> how many items it helps make
        self.received_recipes = defaultdict(set) # learner -> set of (p1, p2)

    def share_memories(self, i_step, agent_states):
        # 1. Map Global Progress
        current_inv_counts = Counter()
        step_new_disco = []
        
        for aid, state in agent_states.items():
            inv = state['inventory']
            current_inv_counts.update(inv)
            
            # Update knowledge from latest memories
            mems = state['memories']
            for m_idx in range(max(0, len(mems)-20), len(mems)):
                p1, p2, res = mems[m_idx]
                if res is not None:
                    if res not in self.global_knowledge:
                        self.global_knowledge[res] = (aid, m_idx, (p1, p2))
                        self.item_centrality[p1] += 1
                        self.item_centrality[p2] += 1
                        step_new_disco.append(res)
                
        # Identify Hubs (Top 10% by centrality)
        hub_items = set()
        if self.item_centrality:
            threshold = np.percentile(list(self.item_centrality.values()), 90)
            hub_items = {item for item, val in self.item_centrality.items() if val >= threshold and val > 1}

        memory_exchanges = {aid: [] for aid in range(self.n_agents)}
        
        # 2. Strategic Distribution
        for l_id, state in agent_states.items():
            l_inv = set(state['inventory'])
            candidates = []
            
            for res, (t_id, m_idx, (p1, p2)) in self.global_knowledge.items():
                if res in l_inv or t_id == l_id:
                    continue
                
                # Check if already shared
                recipe_key = tuple(sorted([str(p1), str(p2)]))
                if recipe_key in self.received_recipes[l_id]:
                    continue

                score = 0
                has_p1 = p1 in l_inv
                has_p2 = p2 in l_inv
                
                # Tier scoring
                if has_p1 and has_p2:
                    score += 50 # Immediate Action
                elif has_p1 or has_p2:
                    score += 20 # Near-Frontier
                
                if res in hub_items:
                    score += 30 # Infrastructure
                
                if res in step_new_disco:
                    score += 15 # Freshness
                
                # Guild Specialization (first 30% of steps)
                if i_step < (self.n_steps * 0.3):
                    # Guilds focus on recipes including their 'native' basics
                    # Guild 0: Air, 1: Water, 2: Earth, 3: Fire (approx)
                    if (i_step % 10) == self.agent_guilds[l_id]:
                        score += 10
                
                if score > 0:
                    candidates.append((score, t_id, m_idx, res in hub_items, (has_p1 or has_p2)))

            # Sort and apply bottleneck
            candidates.sort(key=lambda x: x[0], reverse=True)
            selection = candidates[:20]
            
            for score, t_id, m_idx, is_hub, is_actionable in selection:
                memory_exchanges[l_id].append((t_id, m_idx))
                p1, p2, _ = agent_states[t_id]['memories'][m_idx]
                self.received_recipes[l_id].add(tuple(sorted([str(p1), str(p2)])))
        return memory_exchanges
\end{tcbverbatim}


\subsection{Baseline protocols}
\label{app:baseline_protocols}
\begin{tcbverbatim}[Baseline protocol: asocial (no memory exchange)]{basicstyle=\ttfamily\scriptsize}
class AsocialProtocol:
    def __init__(self, n_agents, n_steps):
        self.n_agents = n_agents
        self.n_steps = n_steps

    def share_memories(self, i_step, agent_states):
        return {aid: [] for aid in agent_states.keys()}
        
\end{tcbverbatim}
\begin{tcbverbatim}[Baseline protocol: graph (players exchange memories with their neighbors in a randomly connected graph (Erdős–Rényi))]{basicstyle=\ttfamily\scriptsize}
class GraphProtocol:
    def __init__(self, n_agents, n_steps, p_conn=0.2, share_prob=0.5, n_memories=10):
        self.n_agents = n_agents
        self.n_steps = n_steps
        self.p_conn = p_conn
        self.share_prob = share_prob
        self.n_memories = n_memories

        # Sample symmetric adjacency matrix
        rand_mat = np.random.rand(n_agents, n_agents)
        adj = (rand_mat < p_conn).astype(int)
        adj = np.triu(adj, 1)
        adj = adj + adj.T
        np.fill_diagonal(adj, 0)
        self.adj = adj

        # Precompute neighbor lists
        self.neighbors = {i: np.where(adj[i])[0].tolist() for i in range(n_agents)}

    def share_memories(self, i_step, agent_states):
        exchanges = {i: [] for i in agent_states}
        total_shared = 0

        for learner in agent_states.keys():
            if np.random.rand() > self.share_prob:
                continue
            neigh = self.neighbors[learner]
            if not neigh:
                continue

            teacher = random.choice(neigh)
            teacher_mems = agent_states[teacher]["memories"]
            if len(teacher_mems) == 0:
                continue

            n = min(self.n_memories, len(teacher_mems))
            mem_ids = np.random.choice(range(len(teacher_mems)), n, replace=False)
            exchanges[learner].extend([(teacher, int(mid)) for mid in mem_ids])
            total_shared += n

        return exchanges
\end{tcbverbatim}
\begin{tcbverbatim}[Baseline protocol: paired (pairs of agents exchange memories)]{basicstyle=\ttfamily\scriptsize}
import numpy as np

class PairedProtocol:

    def __init__(self, n_agents, n_steps):
        self.n_agents = n_agents
        self.n_steps = n_steps
        self.n_memories = 10
        self.share_prob = 0.5
        # create stable pairs
        ids = list(range(n_agents))
        np.random.shuffle(ids)
        self.pairs = {ids[i]: ids[i+1] for i in range(0, n_agents, 2)}
        self.pairs.update({v: k for k, v in self.pairs.items()})

    def share_memories(self, i_step, agent_states):
        exchanges = {aid: [] for aid in agent_states}
        for learner_id in agent_states:
            if np.random.rand() > self.share_prob:
                continue
            teacher_id = self.pairs[learner_id]
            teacher_mems = agent_states[teacher_id]["memories"]
            if not teacher_mems:
                continue
            n = min(len(teacher_mems), self.n_memories)
            chosen = np.random.choice(range(len(teacher_mems)), size=n, replace=False)
            exchanges[learner_id] = [(teacher_id, int(mid)) for mid in chosen]

        return exchanges
\end{tcbverbatim} 
\begin{tcbverbatim}[Baseline protocol: dynamic (paired with additional random visits)]{basicstyle=\ttfamily\scriptsize}
import numpy as np
import random

class DynamicProtocol():
    def __init__(self, n_agents, n_steps):
        self.n_agents = n_agents
        self.n_steps = n_steps
        ids = list(range(n_agents))
        np.random.shuffle(ids)
        self.pairs = {ids[i]: ids[i + 1] for i in range(0, n_agents, 2)}
        self.pairs.update({v: k for k, v in self.pairs.items()})
        self.visit_in_progress = False
        self.visit_time = 0
        self.visit_duration = 20
        self.p_visit = 1
        self.visiting_agent_id = None
        self.visited_agent_ids = []
        self.n_memories = 10
        self.share_prob = 0.5

    def share_memories(self, i_step, agent_states):
        # now there is always a visit (p=1)
        memory_exchanges = {}
        learner_ids = list(agent_states.keys())
        for learner_id in learner_ids:
            if np.random.rand() > self.share_prob:
                continue

            if self.visit_in_progress and learner_id == self.visiting_agent_id:
                # the agent is visiting, so it learns from visited agents
                teacher_ids = self.visited_agent_ids
            elif learner_id in self.visited_agent_ids:
                # the agent is visited, so it learns from its pair and the visiting agent
                teacher_ids = [self.pairs[learner_id], self.visiting_agent_id]
            else:
                if self.pairs[learner_id] == self.visiting_agent_id:
                    # if the pair is on a visit, the agent learns from one
                    continue
                else:
                    # the agent learns from its pair
                    teacher_ids = [self.pairs[learner_id]]
            memory_exchanges[learner_id] = []
            memories = []
            for teacher_id in teacher_ids:
                teacher_state = agent_states[teacher_id]
                n_teacher_memories = len(teacher_state['memories'])
                memories += [(teacher_id, i_mem) for i_mem in range(n_teacher_memories)]
            n_memories = min(len(memories), self.n_memories)  # n_memories to transmit
            memory_ids = np.random.choice(range(len(memories)), size=n_memories, replace=False)
            for mem_id in memory_ids:
                mem = memories[mem_id]
                memory_exchanges[learner_id].append(mem)


        # track visits in progress
        if self.visit_in_progress:
            self.visit_time += 1
            if self.visit_time == self.visit_duration:
                self.visit_in_progress = False
                self.visit_time = 0
                self.visited_agent_ids = []
                self.visiting_agent_id = None

        # start a visit
        if not self.visit_in_progress and np.random.rand() < self.p_visit:
            self.visit_in_progress = True
            self.visit_time = 0
            # sample a visiting agent
            self.visiting_agent_id = int(np.random.choice(learner_ids))
            # sample a pair to visit
            other_agent_ids = [i for i in learner_ids if i not in [self.visiting_agent_id, self.pairs[self.visiting_agent_id]]]
            visited_agent_id = int(random.choice(other_agent_ids))
            self.visited_agent_ids = [visited_agent_id, self.pairs[visited_agent_id]]

        return memory_exchanges
        
\end{tcbverbatim}
\begin{tcbverbatim}[Baseline protocol: stochastic (stochastic, maximal memory exchanges)]{basicstyle=\ttfamily\scriptsize}
import random
import numpy as np

class StochasticProtocol:
    def __init__(self, n_agents, n_steps):
        self.n_agents = n_agents
        self.n_steps = n_steps
        self.n_memories = 20

    def share_memories(self, i_step, agent_states):
        exchanges = {aid: [] for aid in agent_states}
        agent_ids = list(agent_states.keys())

        for learner_id in agent_ids:
            exchanges[learner_id] = []
            teacher_ids = [i for i in agent_ids if i != learner_id]
            for i in range(self.n_memories):
                teacher_id = random.choice(teacher_ids)
                teacher_mems = agent_states[teacher_id]["memories"]
                if not teacher_mems:
                    continue
                chosen = np.random.choice(range(len(teacher_mems)), size=1)
                exchanges[learner_id] += [(teacher_id, int(mid)) for mid in chosen]
        return exchanges
\end{tcbverbatim}

\subsection{Prompts}
\label{app:prompts}
\begin{tcbverbatim}[Prompt of the LLM agent modeling a human player (Gemini 2.0 Flash Lite)]{basicstyle=\ttfamily\scriptsize}
You are playing a discovery game where the goal is to uncover new items by combining existing ones.

# Task Description
At each step, you are given:
- **Personal Inventory**: the items you currently have access to and can combine
- **Personal Memory**: combinations you have already attempted (including failed ones)
- **Social Inventory**: items made available by your peers for the next combination
- **Social Memory**: combinations attempted or discovered by other players (including failed ones)

Your task is to choose a pair of items from your inventory and the social inventory and attempt to combine them. If the combination succeeds, the discovered item will be added to your personal inventory.

The objective of the game is to collectively (you and other player) discover as many different items as possible.

# Rules

1. Only use items that are currently in your inventory or in the social inventory
2. The order of items does not matter ("air + fire" is the same as "fire + air")
3. Combining the same items always gives the same result (deterministic recipes)
4. You can combine an item with itself (e.g., "air + air").

# Response Format

First, think through your reasoning inside:
<reasoning>
... your internal analysis here ...
</reasoning>

Then select two items to combine:
<answer>
item1 + item2
</answer>

Make sure the items are in your inventory or the social inventory.
Use the "+" between the item names.

# Tips

* Keep reasoning short and focused.
* In <reasoning>, always list 2-5 candidate combinations before choosing one.
* For each candidate, explicitly predict the resulting item and explain why, using this format:
  * `a + b $\rightarrow$ c?` --- short explanation based on a real physical process or a clear metaphor.
* Avoid repeating past combinations found in personal or social memories.
* Make sure to try combination you think will result in new items you don't have yet.
* Prefer recently discovered items that haven't been combined much yet, but don't rely only on the most recent ones.
* Items in the social inventory are only available right now, use them now if you think they can be useful

# Example

<reasoning>
I have wood, stone, water, and heat, and the social inventory gives me metal and cloth for this turn only.
Here are possible combinations and the new items they might produce:
 * wood + metal $\rightarrow$ tool handle or basic tool? --- many tools are made by attaching metal to wood, it would be very useful for later combinations
 * wood + water $\rightarrow$ boat? --- this makes sense, and could be useful for later combinations
 * water + cloth $\rightarrow$ swimsuit? --- realistic, but not very useful for future discoveries
Metal is likely to be especially valuable, and it disappears after this turn so I'll try to use it.
Between the options, **wood + metal** is the most plausible way to obtain a general-purpose tool that could unlock many future combinations.
I'll try **wood + metal**.
</reasoning>

<answer>
wood + metal
</answer>

# Current state:

**Personal Inventory** (in order of discovery):
{inventory}

**Personal Memory**:
{memory}

**Social Inventory**:
{social_inventory}

**Social Memory**:
{social_memory}

\end{tcbverbatim}
\begin{tcbverbatim}[Prompt of the LLM involved in the evolurionary search algorithm (Gemini 3.0 Flash)]{basicstyle=\ttfamily\scriptsize}

# Transmission Protocol Design

Design a transmission protocol that organizes information sharing or coordination in a population of agents working toward a shared discovery task.

## Environment Overview

Agents play a combinatorial exploration game with 720 elements. Each agent starts with 4 basics (air, water, earth, fire) and, each step, combines two elements to try to discover new ones. Combinations are deterministic (produce a fixed element or nothing). At each step:
1) Sharing: the protocol decides which past recipes (memories) each agent receives from others.
2) Borrowing: agents may temporarily use elements from received recipes for this step only (two combined elements, and the possible resulting element).
3) Action: each agent performs one combination.

There are n_agents agents and the game runs across n_steps steps.

## Objective

Maximize **global coverage** (fit_item_coverage_global): the number of unique elements discovered across all agents.
You should design a protocol that helps the group as a whole perform better through structured communication. Consider how agents decide **when** to communicate, **what** to share, and **who** they share with. Your design can be static or adaptive --- parameters may change over time if useful.

## Protocol Interface (must implement)

Define a class `TransmissionProtocol` with:
- `__init__(self, n_agents, n_steps)` --- initialize any state here.
- `share_memories(self, i_step, agent_states)` -> dict: learner_id -> list[(teacher_id, memory_id)]
  - `agent_states[aid] = {'inventory': tuple(item), 'memories': tuple((item1, item2, result))}` with `result=None` if failed
- `get_logs(self)` -> `list[dict(metric_name:str, metric_description:str, metric_value:float)]`

**Notes**
- Transmission is optional: sometimes limited sharing can be beneficial.
- A communication bottleneck prevents players from receiving more than 20 memories at each step. If more are shared, they will be sub-sampled randomly.
- Clearly explain how your design improves group outcomes relative to simpler baselines.
- Use logging (via `get_logs`) to record internal signals that show whether your mechanisms produce intended effects. Logs will help you refine protocols later.


## Execution misc
* Allowed imports include: `numpy as np`, `sklearn.neighbors`, `sklearn.metrics`, `random`, `collections`, `statistics`. Avoid other imports.
* You cannot access agents' internal attributes.
* You cannot read or write files on disk.
* Do not print; record quantities through `get_logs`.
* Each simulation run must complete in under 20 seconds.


## Output format (strict)
Structure your response **exactly** as:

<reasoning>
[Reflect on what the protocol is trying to optimize and what might limit current designs.
Identify specific communication or coordination problems (e.g., inefficiency, over-sharing, lack of structure).
Propose several possible strategies and explain which one you choose and why.
If other solutions are provided, analyze them and learn from their logs, identify their strengths and limitations before designing your own solution.]
</reasoning>

<description>
[Describe in detail how your protocol works: how agents decide when and what to share, and how information flows.
Define any key parameters or thresholds explicitly in `__init__`.
If your design involves randomness or adaptation, explain its purpose.
Summarize why this approach should improve group performance.]
</description>

<logging>
[Describe which quantities you will record in `get_logs()` to verify the behavior of your protocol.
Include general indicators of communication intensity, information overlap, or variation among agents.
Explain briefly how these signals can be used to assess the protocol's effects and can guide future improvements.]
</logging>

<code>
```python
class TransmissionProtocol:
    def __init__(self, n_agents, n_steps):
        # Initialize any state variables
        pass
    def share_memories(self, i_step, agent_states):
        # Your implementation here
        # return {learner_id: [(teacher_id, memory_id), ...], ...}
        return memory_exchanges
    def get_logs(self):
        # return list of dicts with metric_name, metric_description, metric_value
        return list_metric_dicts
    # helper methods (optional)
```
</code>


## Approaches that have been explored
Here are successful transmission protocols that have already been tried:
{examples}


## Statistics of protocols discovered so far

Average protocol: fit_item_coverage_global={mean\_fitness}
Best protocol: fit_item_coverage_global={best\_fitness}

## Instruction

{if mutation mode == diversify}
Review the approaches above and propose a new protocol that takes a fundamentally different direction than all of them. Explore a novel communication or coordination strategy that has not yet been tried.

{elif mutation mode == improve}
Study the parent solution, identify its weaknesses, and propose small, targeted adjustments that preserve its strengths while improving its coordination or efficiency. You may draw inspiration from the alternative approaches.

In <reasoning>, analyze how previous solutions organized information flow, identify their main limitations, and propose a concrete way to improve or rethink this organization.
Remember to use the complete format with each of reasoning, description, logging, and code, using xml tags.

\end{tcbverbatim}

\end{document}